\documentclass[journal]{IEEEtran}
\usepackage{graphicx}
\usepackage{bm} 
\usepackage{multirow}
\usepackage{colortbl}
\newcommand{\vct}[1]{\boldsymbol{#1}} 
\newcommand{\mat}[1]{\boldsymbol{#1}} 
\newcounter{example}

\newcommand{\ignore}[1]{}
\begin{document}
\title{From Two-Class Linear Discriminant Analysis to \\
Interpretable Multilayer Perceptron Design}

\author{Ruiyuan~Lin, Zhiruo~Zhou, Suya~You, Raghuveer~Rao and C.-C.~Jay
Kuo,~\IEEEmembership{Fellow,~IEEE}
\thanks{Ruiyuan~Lin, Zhiruo~Zhou and C.-C.~Jay~Kuo are with Ming Hsieh Department
of Electrical and Computer Engineering, University of Southern
California, 3740 McClintock Avenue, Los Angeles, USA,
emails:\{ruiyuanl,zhiruozh,jckuo\}@usc.edu}
\thanks{Suya You and Raghuveer Rao are with Army Research Laboratory,
Adelphi, Maryland, USA, emails:\{suya.you.civ,raghuveer.m.rao.civ\}@mail.mil}%
}%

\maketitle

\begin{abstract}

A closed-form solution exists in two-class linear discriminant analysis
(LDA), which discriminates two Gaussian-distributed classes in a
multi-dimensional feature space. In this work, we interpret the
multilayer perceptron (MLP) as a generalization of a two-class LDA
system so that it can handle an input composed by multiple Gaussian
modalities belonging to multiple classes. Besides input layer $l_{in}$
and output layer $l_{out}$, the MLP of interest consists of two
intermediate layers, $l_1$ and $l_2$.  We propose a feedforward design
that has three stages: 1) from $l_{in}$ to $l_1$: half-space
partitionings accomplished by multiple parallel LDAs, 2) from $l_1$ to
$l_2$: subspace isolation where one Gaussian modality is represented by
one neuron, 3) from $l_2$ to $l_{out}$: class-wise subspace mergence,
where each Gaussian modality is connected to its target class. Through
this process, we present an automatic MLP design that can specify the
network architecture (i.e., the layer number and the neuron number at a
layer) and all filter weights in a feedforward one-pass fashion.  This
design can be generalized to an arbitrary distribution by leveraging the
Gaussian mixture model (GMM). Experiments are conducted to compare the
performance of the traditional backpropagation-based MLP (BP-MLP) and
the new feedforward MLP (FF-MLP). 

\end{abstract}


\IEEEpeerreviewmaketitle

\section{Introduction}\label{sec:introduction}

The multilayer perceptron (MLP), proposed by Rosenblatt in 1958
\cite{rosenblatt1958perceptron}, has a history of more than 60 years.
However, while there are instances of theoretical investigations into
why the MLP works (e.g., the classic articles of Cybenko
\cite{cybenko1989approximation}, and Hornik, Stinchcombe and White
\cite{hornik1989multilayer}), the majority of the efforts have focused on
applications such as speech recognition \cite{ahad2002speech}, economic
time series \cite{devadoss2013forecasting}, image processing
\cite{sivakumar1993image}, and many others. 

One-hidden-layer MLPs with suitable activation functions are shown to be universal approximators  \cite{cybenko1989approximation,hornik1989multilayer,lu2017expressive}. Yet, this only shows the existence but does not provide guideline in network design \cite{csaji2001approximation}.
Sometimes, deeper networks could be more efficient than shallow wider networks.  The MLP design
remains to be an open problem. In practice, 
trials and errors are made in determining the
layer number and the neuron number of each layer. The process of hyper
parameter finetuning is time consuming. We attempt to address these two
problems simultaneously in this work: MLP theory and automatic MLP
network design. 

For MLP theory, we will examine the MLP from a brand new angle. That is,
we view an MLP as a generalization form of the classical two-class
linear discriminant analysis (LDA).  The input to a two-class LDA system
is two Gaussian distributed sources, and the output is the predicted class.  The
two-class LDA is valuable since it has a closed-form solution. Yet, its
applicability is limited due to the severe constraint in the problem
set-up. It is desired to generalize the LDA so that an arbitrary
combination of multimodal Gaussian sources represented by multiple
object classes can be handled.  If there exists such a link between MLP
and two-class LDA, analytical results of the two-class LDA can be
leveraged for the understanding and design of the MLP. The
generalization is possible due to the following observations. 
\vspace{-0ex}
\begin{itemize}
\item The first MLP layer splits the input space with multiple partitioning
hyperplanes. We can also generate multiple partitioning hyperplanes with
multiple two-class LDA systems running in parallel.
\item With specially designed weights, each neuron in the second MLP
layer can activate one of the regions formed by the first layer
hyperplanes. 
\item A sign confusion problem arises when two MLP layers are in
cascade. This problem is solved by applying the rectified linear unit
(ReLU) operation to the output of each layer. 
\end{itemize}

In this paper, we first make an explicit connection between the
two-class LDA and the MLP design. Then, we propose a general MLP
architecture that contains input layer $l_{in}$, output layer $l_{out}$,
two intermediate layers, $l_1$ and $l_2$.  Our MLP design consists of
three stages:
\begin{itemize}
\item Stage 1 (from input layer $l_{in}$ to $l_1$): Partition the whole
input space flexibly into a few half-subspace pairs, where the
intersection of half-subspaces yields many regions of interest. 
\item Stage 2 (from intermediate layer $l_1$ to $l_2$): Isolate each region of
interest from others in the input space. 
\item Stage 3 (from intermediate layer $l_2$ to output layer $l_{out}$):
Connect each region of interest to its associated class. 
\end{itemize}
The proposed design can determine the MLP architecture and weights of
all links in a feedforward one-pass manner without trial and error.  No
backpropagation is needed in network training. 

In contrast with traditional MLPs that are trained based on end-to-end
optimization through backpropagation (BP), it is proper to call our new
design the feedforward MLP (FF-MLP) and traditional ones the
backpropagation MLP (BP-MLP). Intermediate layers are not hidden but
explicit in FF-MLP.  Experiments are conducted to compare the
performance of FF-MLPs and BP-MLPs. The advantages of FF-MLPs over
BP-MLPs are obvious in many areas, including faster design time and
training time. 

The rest of the paper is organized as follows.  The relationship between the
two-class LDA and the MLP is described in Sec.  \ref{sec:lda}. A
systematic design of an interpretable MLP in a one-pass feedforward
manner is presented in Sec. \ref{sec:design}. Several MLP design
examples are given in Sec.  \ref{sec:examples}.  Observations on the
BP-MLP behavior are stated in Sec.  \ref{sec:insights}.  We compare the
performance of BP-MLP and FF-MLP by experiments in Sec.
\ref{sec:experiments}.  Comments on related previous work are made in
Sec. \ref{sec:comments}.  Finally, concluding remarks and future
research directions are given in Sec.  \ref{sec:conclusion}. 

\section{From Two-Class LDA to MLP}\label{sec:lda}

\subsection{Two-Class LDA}\label{subsec:tclda}

\begin{figure}[t]
\centering
\begin{minipage}[b]{0.8\linewidth}
\centerline{\includegraphics[width=1.0\linewidth]{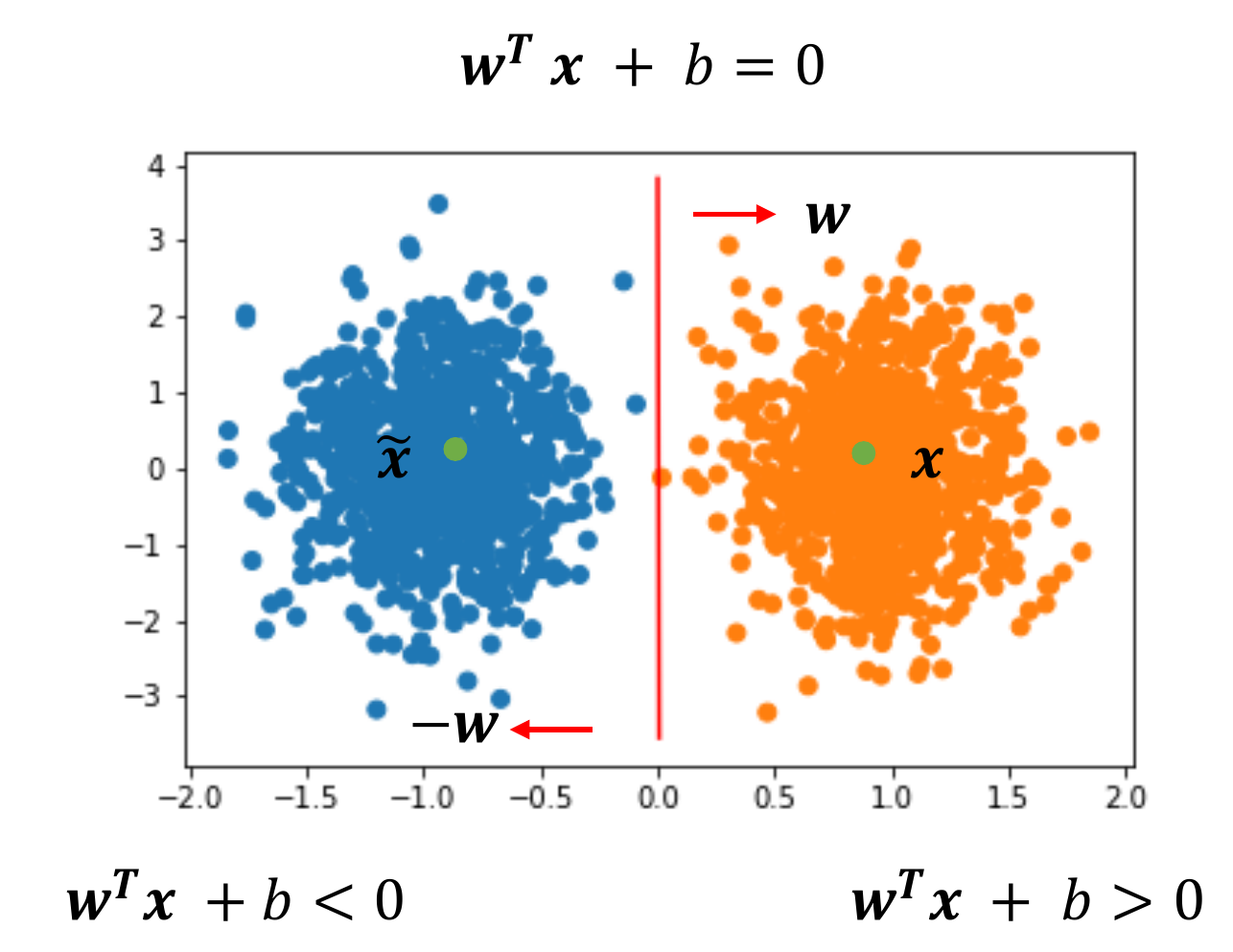}}
\centerline{(a)}
\end{minipage}
\begin{minipage}[b]{0.9\linewidth}
\centering
\centerline{\includegraphics[width=1.0\linewidth]{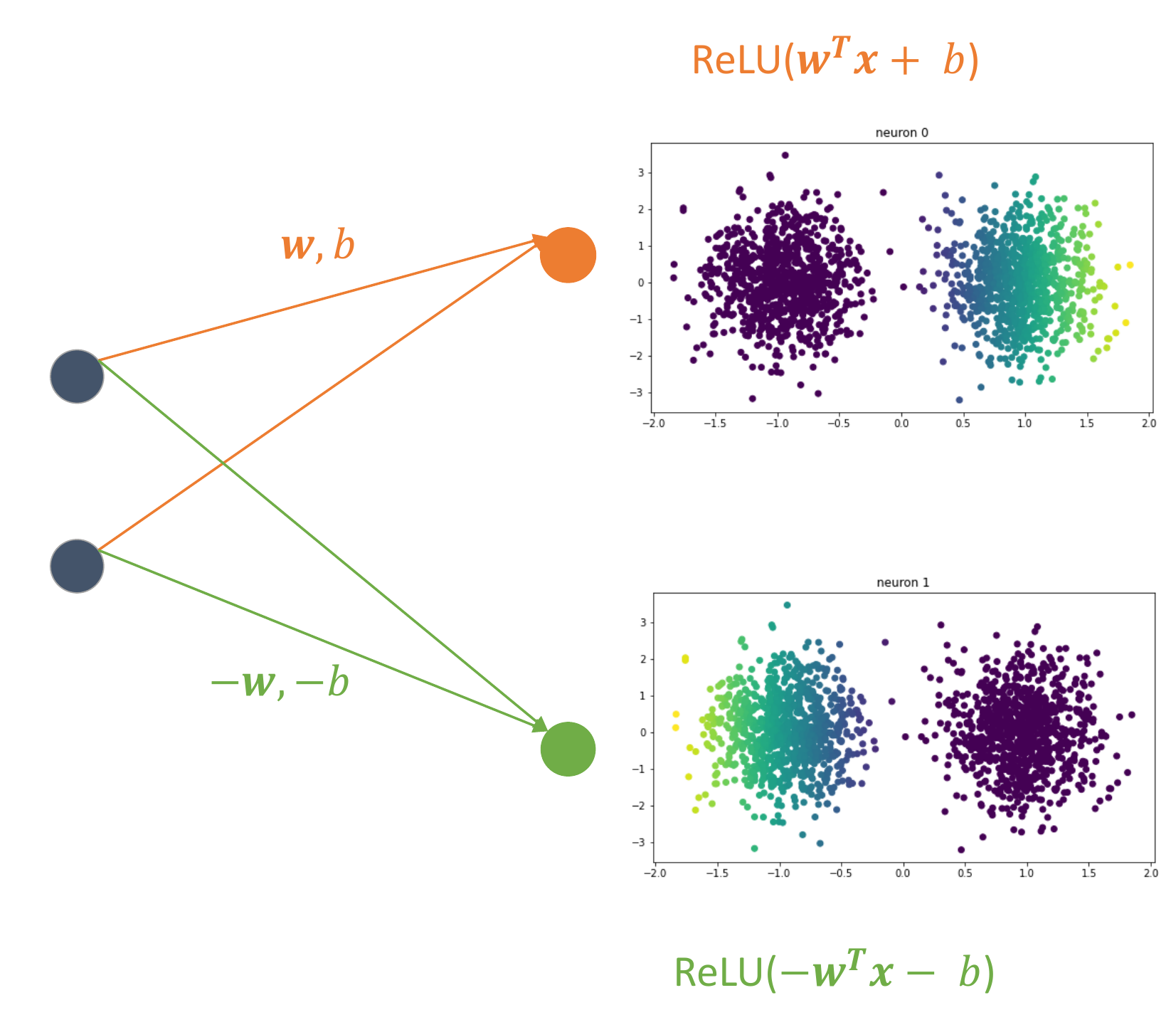}}
\centerline{(b)}
\end{minipage}
\caption{Illustration of (a) two Gaussian blobs separated by a line in a
2D plane, and (b) a one-layer two-neuron perceptron system.}\label{fig:2gaussian}
\end{figure}

Without loss of generality, we consider two-dimensional (2D) random
vectors, denoted by ${\bf x} \in R^2$, as the input for ease of
visualization. They are samples from two classes, denoted by $C_1$ and
$C_2$, each of which is a Gaussian-distributed function. The two
distributions can be expressed as $\mathcal{N}(\mu_1,\Sigma_1)$ and
$\mathcal{N}(\mu_2,\Sigma_2)$, where $\vct{\mu_1}$ and $\vct{\mu_2}$ are
their mean vectors and $\mat{\Sigma_1}$ and $\mat{\Sigma_2}$ their covariance
matrices, respectively. Fig.  \ref{fig:2gaussian}(a) shows an example of
two Gaussian distributions, where the blue and orange dots are samples
from classes $C_1$ and $C_2$, respectively. Each Gaussian-distributed
modality is called a Gaussian blob. 

A linear classifier can be used to separate the two blobs, known as the
two-class linear discriminant analysis (LDA), in closed form.  LDA assumes homoscedasticity, that is, the covariances of different classes are identical: $\mat{\Sigma_1} = \mat{\Sigma_2} =  \mat{\Sigma}$. 
In this case, the decision boundary can be formulated into the form of \cite{scikit-learn}\cite{friedman2001elements}
\begin{equation}\label{eq:separating}
\vct{w}^T\vct{x} + b = 0,
\end{equation}
where 
\begin{equation}\label{eq:normal}
{\bf w}=(w_1,w_2)^T= \mat{\Sigma}^{-1} (\vct{\mu_1}-\vct{\mu_2}).
\end{equation}
\begin{equation}\label{eq:bias}
b = \frac{1}{2}\vct{\mu_2}^T \mat{\Sigma}^{-1} \vct{\mu_2}-\frac{1}{2}\vct{\mu_1}^T \mat{\Sigma}^{-1} \vct{\mu_1} + \log{\frac{p}{1-p}}, 
\end{equation}
where $p=P(y=1)$.
Then, sample ${\bf x}$ is classified to class $C_1$ if $\vct{w}^T\vct{x} + b$ evaluates positive. Otherwise, sample ${\bf x}$ is classified to class $C_2$.

\subsection{One-Layer Two-Neuron Perceptron}\label{subsec:12perceptron}

We can convert the LDA in Sec. \ref{subsec:tclda} to a one-layer
two-neuron perceptron system as shown in Fig. \ref{fig:2gaussian}(b).
The input consists of two nodes, denoting the first and the second
dimensions of random vector ${\bf x}$.  The output consists of two
neurons in orange and green, respectively.  The two orange links have weight $w_1$ and $w_2$ 
that can be determined based on Eq.  (\ref{eq:normal}). The bias $b$ for the orange node can be obtained based on Eq.  (\ref{eq:bias}).
Similarly, the two green links have weight $-w_1$ and $-w_2$ and the green node has bias $-b$. 
The rectified linear unit (ReLU), defined as
\begin{equation}\label{eq:relu}
\mbox{ReLU}(y)=\max(0,y),
\end{equation}
is chosen to be the activation function in the neuron. The activated
responses of the two neurons are shown in the right part of Fig.
\ref{fig:2gaussian}(b). The left (or right) dark purple region in the top (or bottom) subfigure means zero responses. Responses are non-zero in the
other half. We see more positive values as moving further to the right
(or left). 

One may argue that there is no need to have two neurons in Fig.
\ref{fig:2gaussian}(b).  One neuron is sufficient in making correct
decision. Although this could be true, some information of the two-class
LDA system is lost in a one-neuron system.  The magnitude of the
response value for samples in the left region is all equal to zero if
only the orange output node exists. This degrades the classification
performance. In contrast, by keeping both orange and green nodes, we can
preserve the same amount of information as that in the two-class LDA. 

One may also argue that there is no need to use the ReLU activation
function in this example. Yet, ReLU activation is essential when the
one-layer perceptron is generalized to a multi-layer perceptron as
explained below.  We use $\tilde{{\bf x}}$ to denote the mirror (or the
reflection) of ${\bf x}$ against the decision line in Fig.
\ref{fig:2gaussian}(a) as illustrated by the pair of green dots.
Clearly, their responses are of the opposite sign. The neuron response in the next stage is the sum of multiple response-weight
products. Then, one cannot distinguish the following two cases:
\vspace{-0ex}
\begin{itemize}
\item a positive response multiplied by a positive weight,
\item a negative response multiplied by a negative weight;
\end{itemize}
since both contribute to the output positively. Similarly,
one cannot distinguish the following two cases, either:
\vspace{-0ex}
\begin{itemize}
\item a positive response multiplied by a negative weight,
\item a negative response multiplied by a positive weight;
\end{itemize}
since both contribute to the output negatively.  As a
result, the roles of ${\bf x}$ and its mirror $\tilde{{\bf x}}$ are
mixed together.  The sign confusion problem was first pointed out in
\cite{kuo2016understanding}. This problem can be resolved by the ReLU
nonlinear activation. 

\subsection{Need of Multilayer Perceptron}\label{subsec:mperceptron}

\begin{figure}[t]
\centering
\begin{minipage}[b]{0.9\linewidth}
\centerline{\includegraphics[width=0.9\linewidth]{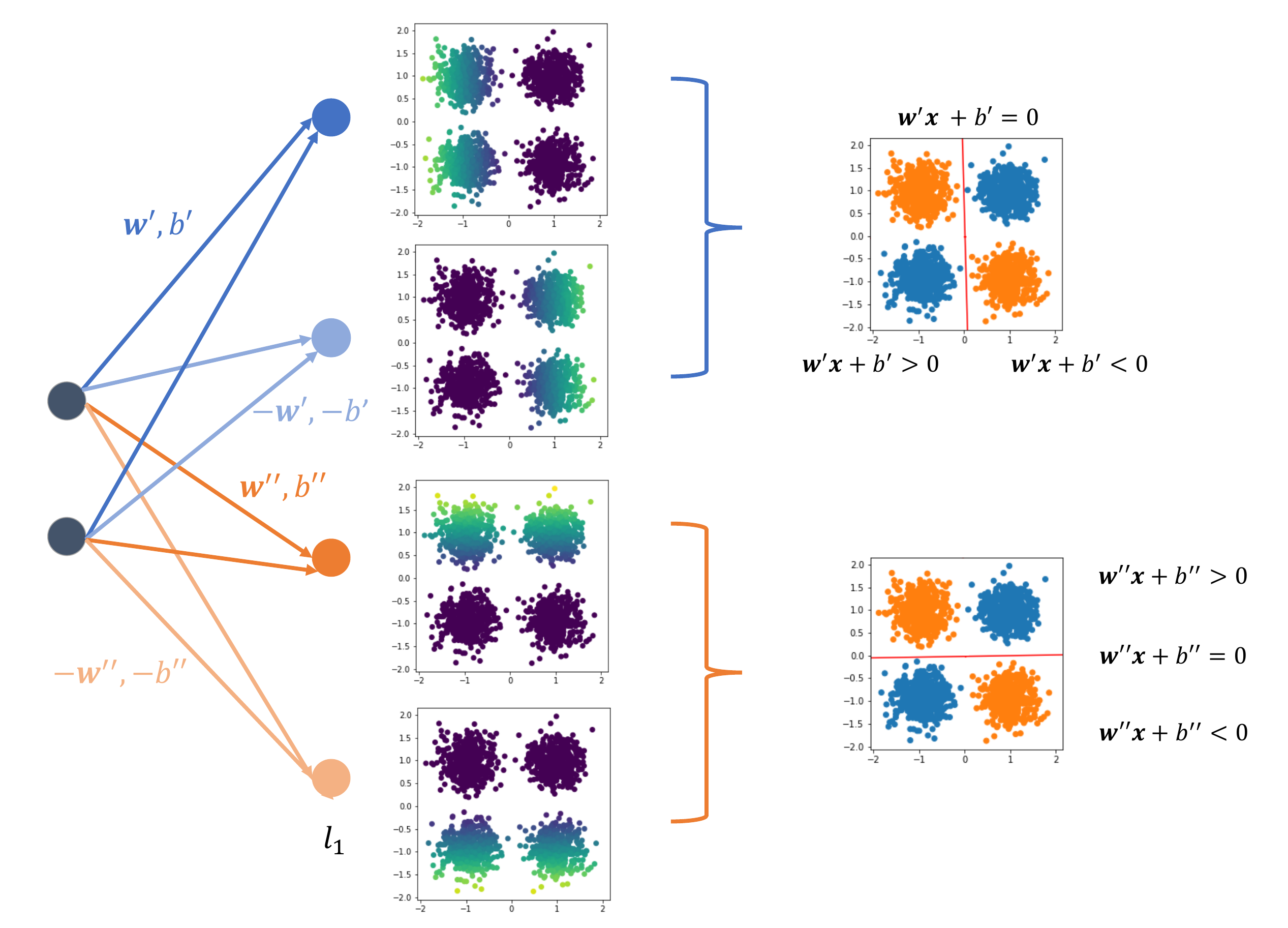}}
\centerline{(a)}
\end{minipage}
\begin{minipage}[b]{0.9\linewidth}
\centerline{\includegraphics[width=0.8\linewidth]{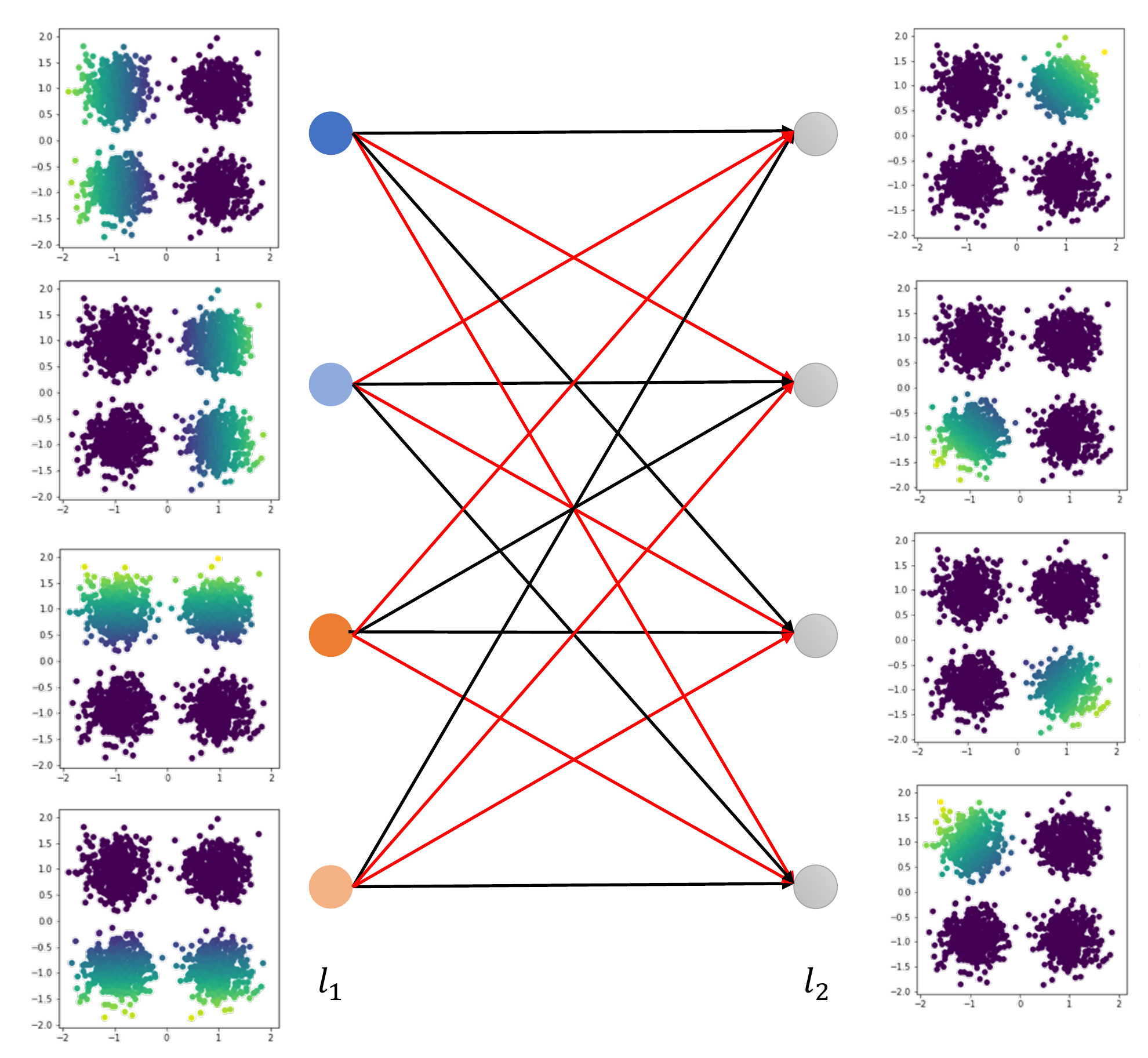}}
\centerline{(b)}
\end{minipage}
\begin{minipage}[b]{0.9\linewidth}
\centerline{\includegraphics[width=0.8\linewidth]{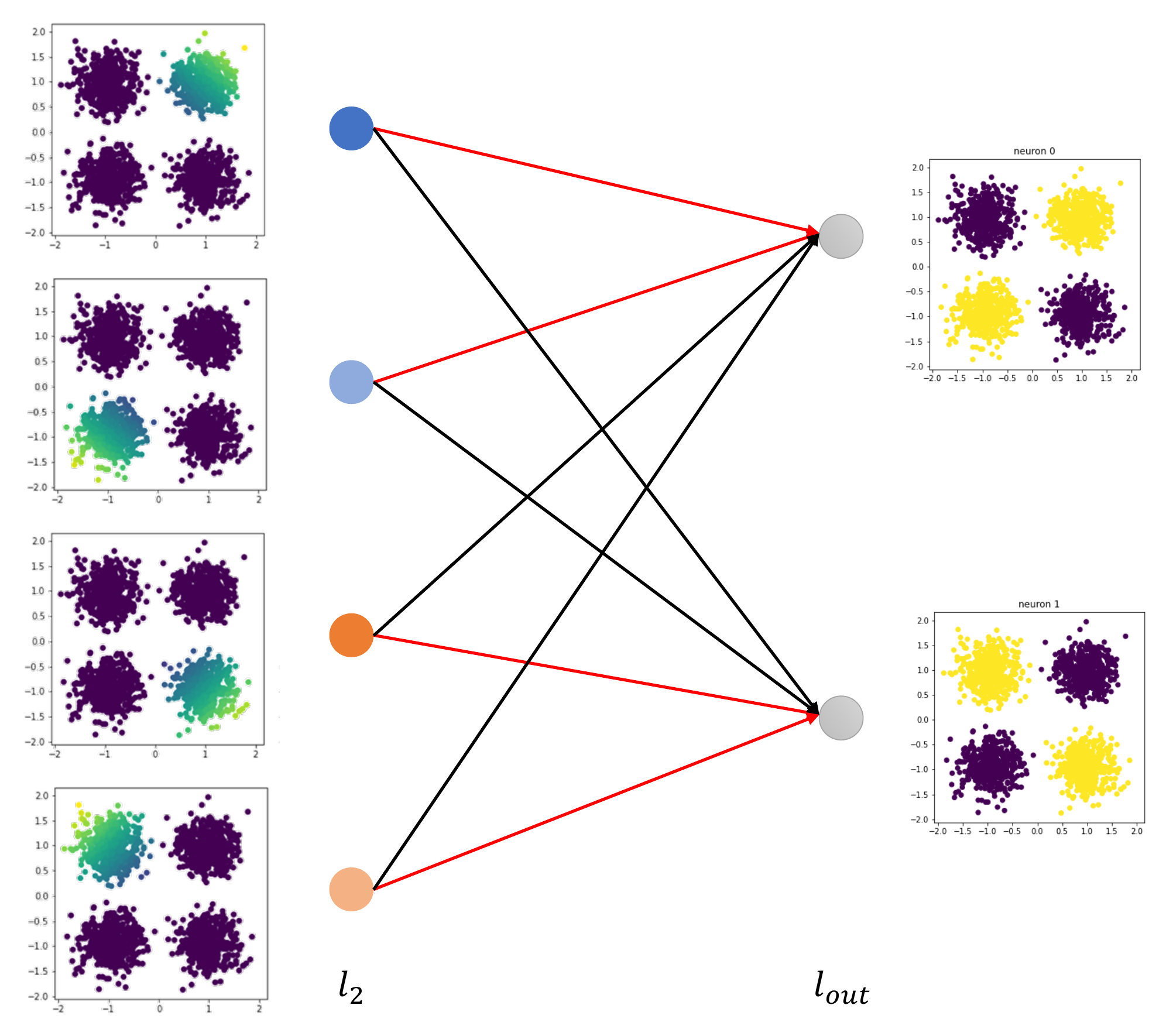}}
\centerline{(c)}
\end{minipage}
\caption{MLP design for Example 1 (XOR): (a) weights between input layer
$l_{in}$ and layer $l_{1}$, (b) weights between layer $l_{1}$ and layer
$l_{2}$, where red links represent weight of 1 and black links represent
weights $-P$, where $P$ is assumed to be a sufficiently large positive number, (c)
weights between $l_{2}$ and $l_{out}$, where red links represent weight
of 1 and black links represent weight of 0.}\label{fig:xor_example}
\end{figure}

Samples from multiple classes cannot be separated by one linear 
decision boundary in general. One simple example is given below.

\noindent {\bf Example 1 (XOR)}. The sample distribution of the XOR
pattern is given in Fig.  \ref{fig:xor_example}(a). It has four Gaussian
blobs belonging to two classes. Each class corresponds to the
``exclusive-OR" output of coordinates' signs of inputs of a blob.  That
is, Gaussian blobs located in the quadrant where x-axis and y-axis have
the same sign belong to class 0. Otherwise, they belong to class 1. The
MLP has $l_{in}$, $l_1$, $l_2$ and $l_{out}$ four layers and three
stages of links. We will examine the design stage by stage in a
feedforward manner. 

\subsubsection{Stage 1 (from $l_{in}$ to $l_{1}$)} Two partitioning
lines are needed to separate four Gaussian blobs -- one vertical and one
horizontal ones as shown in the figure. Based on the discussion in Sec.
\ref{subsec:tclda}, we can determine two weight vectors, ${\bf w'}$ and
${\bf w''}$, which are vectors orthogonal to the vertical and
horizontal lines, respectively. In other words, we have two LDA systems
that run in parallel between the input layer and the first intermediate
layer\footnote{We do not use the term ``hidden" but ``intermediate"
since all middle layers in our feedforward design are explicit rather
than implicit.} of the MLP as shown in Fig.  \ref{fig:xor_example}(a).
Layer $l_1$ has four neurons.  They are partitioned into two pairs of
similar color: blue and light blue as one pair and orange and light
orange as another pair.  The responses of these four neurons are shown
in the figure. By appearance, the dimension goes from two to four from
$l_{in}$ to $l_1$.  Actually, each pair of nodes offers a complementary
representation in one dimension. For example, the blue and light blue
nodes cover the negative and the positive regions of the horizontal
axis, respectively. 

\subsubsection{Stage 2 (from $l_{1}$ to $l_{2}$)} To separate four
Gaussian blobs in Example 1 completely, we need layer $l_2$ as shown in
Fig.  \ref{fig:xor_example}(b). The objective is to have one single
Gaussian blob represented by one neuron in layer $l_2$.  We use the blob
in the first quadrant as an example. It is the top node in layer $l_2$.
The light blue and the orange nodes have nonzero responses in this region, and
we can set their weights to 1 (in red).  There is however a side effect
- undesired responses in the second and the fourth quadrants are brought
in as well.  The side effect can be removed by subtracting responses
from the blue and the light orange nodes. In Fig.
\ref{fig:xor_example}(b), we use red and black links to represent weight
of $1$ and $-P$, where $P$ is assumed to be a sufficiently large positive number\footnote{We set $P$ to 1000 in our experiments.},
respectively. With this assignment, we can preserve responses in the
first quadrant and make responses in the other three quadrants negative.
Since negative responses are clipped to zero by ReLU, we obtain the
desired response. 

\subsubsection{Stage 3 (from $l_{2}$ to $l_{out}$)} The top two nodes
belong to one class and the bottom two nodes belong to another class in
this example.  The weight of a link is set to one if it connects a
Gaussian blob to its class. Otherwise, it is set to zero. All bias terms
are zero. 

\section{Design of Feedforward MLP (FF-MLP)}\label{sec:design}

In this section, we generalize the MLP design for Example 1 so that it
can handle an input composed by multiple Gaussian modalities belonging
to multiple classes.  The feedforward design procedure is stated in Sec.
\ref{subsec:pipeline}. Pruning of partitioning hyperplanes is discussed
in Sec.  \ref{subsec:pruning}. Finally, the designed FF-MLP architecture
is summarized in Sec. \ref{subsec:FF_MLP_summary}.

\subsection{Feedfoward Design Procedure}\label{subsec:pipeline}

\begin{figure}[t]
\centerline{\includegraphics[width=1.0\linewidth]{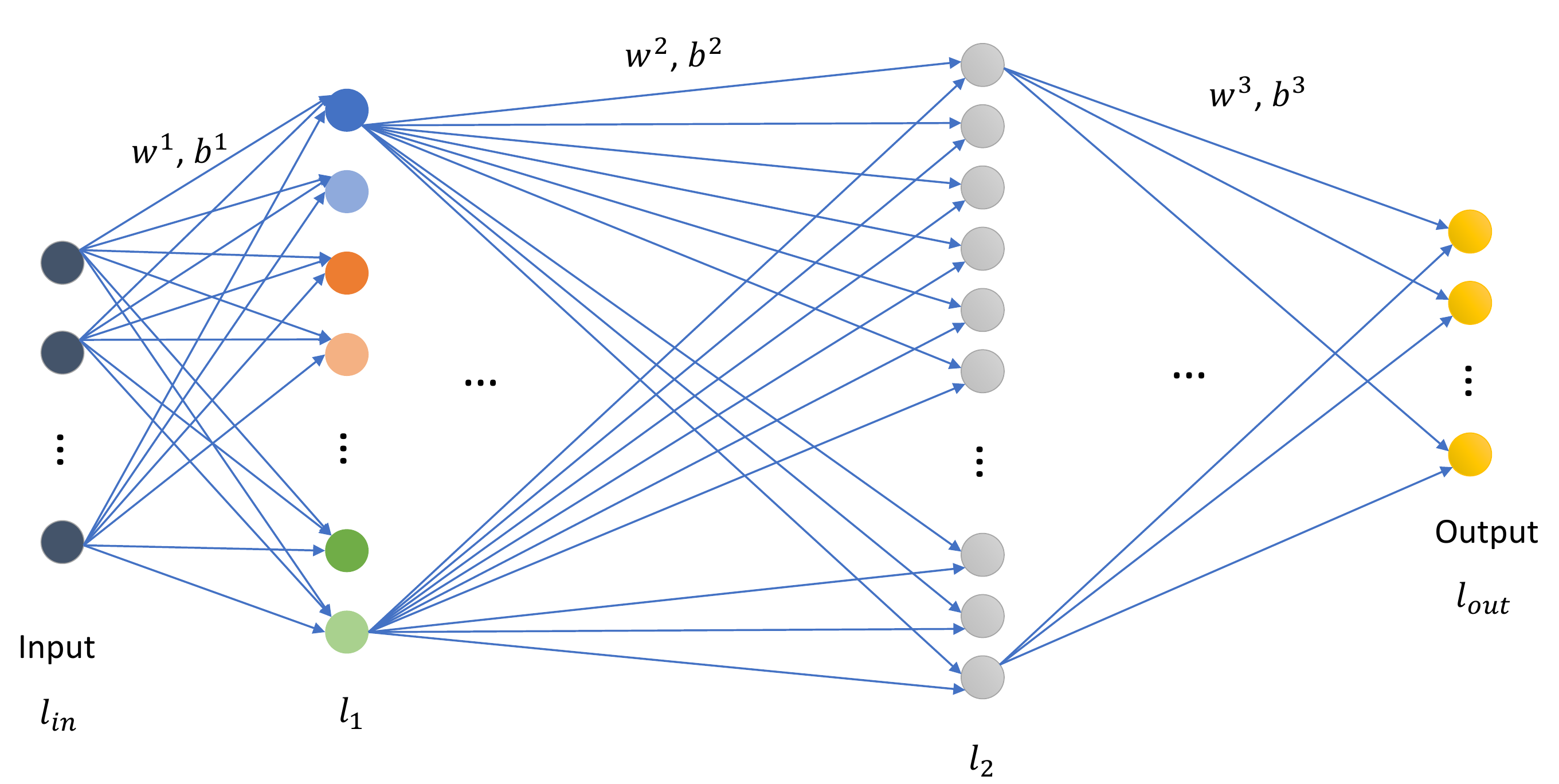}}
\caption{Proposed MLP design: the MLP network with $l_1$ and $l_2$ two intermediate layers, where neurons in layer $l_1$ are drawn in pairs (e.g., blue and light blue nodes) representing two sides of a decision boundary and where each neuron in layer $l_2$ represents an isolated region of
interest.}\label{fig:design}
\end{figure}

We examine an $N$-dimensional sample space formed by $G$ Gaussian blobs
belonging to $C$ classes, where $G \ge C$. The MLP architecture is shown in Fig.
\ref{fig:design}. Again, it has layers $l_{in}$, $l_{1}$, $l_{2}$ and
$l_{out}$.  Their neuron numbers are denoted by $D_{in}$, $D_{1}$,
$D_{2}$ and $D_{out}$, respectively. Clearly, we have
\begin{equation}\label{eq:dim1}
D_{in}=N, \quad D_{out}=C.
\end{equation}
We will show that
\begin{equation}\label{eq:dim2}
D_{1}\le 2 \left( \begin{array}{c}
G \\
2 \end{array}
\right), \quad G \le D_{2} \le 2^{G(G-1)/2}.
\end{equation}
We examine the following three stages one more time but in a more general
setting. 

\subsubsection{Stage 1 (from $l_{in}$ to $l_{1}$) - Half-Space Partitioning} When the input contains $G$ Gaussian blobs of identical covariances, we can select
any two to form a pair and use a two-class LDA to separate them.  Since
there are $L=C^G_2=G(G-1)/2$ pairs, we can run $L$ LDA systems in
parallel and, as a result, the first intermediate layer has $2L$
neurons. This is an upper bound since some partitioning hyperplanes can
be pruned sometimes.  Each LDA system corresponds to an
$(N-1)$-dimensional hyperplane that partitions the $N$-dimensional input
space into two half-spaces represented by a pair of neurons. The weights
of the incident links are normal vectors of the hyperplane of the
opposite signs. The bias term can be also determined analytically.
Interpretation of MLPs as separating hyper-planes is not new (see
discussion on previous work in Sec. \ref{subsec:partitioning}).

\subsubsection{Stage 2 (from $l_{1}$ to $l_{2}$) - Subspace Isolation}
The objective of Stage 2 is to isolate each Gaussian blob in one
subspace represented by one or more neurons in the second intermediate
layer.  As a result, we need $G$ or more neurons to represent $G$
Gaussian blobs.  By isolation, we mean that responses of the Gaussian
blob of interest are preserved while those of other Gaussian blobs are
either zero or negative. Then, after ReLU, only one Gaussian blob of
positive responses is preserved (or activated) while others are clipped
to zero (or deactivated).  We showed an example to isolate a Gaussian
blob in Example 1. This process can be stated precisely below.  

We denote the set of $L$ partitioning hyperplanes by $H_1, H_2, \cdots,
H_L$. Hyperplane $H_l$ divides the whole input space into two
half-spaces $S_{l,0}$ and $S_{l,1}$. Since a neuron of layer $l_1$
represents a half-space, we can use $S_{l,0}$ and $S_{l,1}$ to label the
pair of neurons that supports hyperplane $H_l$. The intersection of $L$
half-spaces yields a region denoted by $R$, which is represented by
binary sequence ${\bf s}(R) = ``c_1, c_2, \cdots , c_L"$ of length $L$
if it lies in half-space $S_{l,c_l}$, $l=1, 2, \cdots, L$.  There are at
most $2^L=2^{G(G-1)/2}$ partitioned regions, and each of them is
represented by binary sequence ${\bf s}(R)$ in layer $l_2$. We assign
weight one to the link from neuron $S_{l,c_l}$ in layer $l_1$ to neuron
${\bf s}(R) = ``c_1, c_2, \cdots , c_L"$ in layer $l_2$, and weight $-P$
to the link from neuron $S_{l,\bar{c_l}}$, where $\bar{c_l}$ is the
logical complement of $c_l$, to the same neuron in layer $l_2$. 

The upper bound of $D_2$ given in Eq. (\ref{eq:dim2}) is a direct
consequence of the above derivation. However, we should point out that
this bound is actually not tight. There is a tighter upper bound for $D_{2}$, which 
is\footnote{The Steiner-Schl\"{a}fli Theorem (1850), as cited in https://www.math.miami.edu/$\sim$armstrong/309sum19/309sum19notes.pdf, p. 21}
\begin{equation}
D_2 \le \sum_{i=0}^{N} 
\left( 
\begin{array}{c} L \\
i \end{array}
\right).
\end{equation}

\subsubsection{Stage 3 (from $l_{2}$ to $l_{out}$) - Class-wise Subspace
Mergence} Each neuron in layer $l_{2}$ represents one Gaussian blob, and
it has $C$ outgoing links. Only one of them has the weight equal to one
while others have the weight equal to zero since it only belongs to one
class. Thus, we connect a Gaussian blob to its target class\footnote{In our experiments, we determine the target class of each region using the majority class in that region.} with weight
``one" and delete it from other classes with weight ``0". 

Since our MLP design does not require end-to-end optimization, no
backprogagation is needed in the training. It is interpretable and its
weight assignment is done in a feedforward manner. To differentiate the
traditional MLP based on backpropagation optimization, we name the
traditional one ``BP-MLP" and ours ``FF-MLP". FF-MLP demands the
knowledge of sample distributions, which are provided by training
samples. 

For general sample distributions, we can approximate the distribution of
each class by a Gaussian mixture model (GMM). 
Then, we can apply the
same technique as developed before.\footnote{In our implementation, we first estimate the GMM parameters using the training data. Then, we use the GMM to generate samples for LDAs.} The number of mixture components is a
hyper-parameter in our design. It affects the quality of the
approximation. When the number is too small, it may not represent the
underlying distribution well. When the number is too large, it may
increase computation time and network complexity. More discussion on 
this topic is given in Example 4.

It is worth noting that the Gaussian blobs obtained by this method are not guaranteed to have the same covariances. Since we perform GMM\footnote{In our experiments, we allow different covariance matrices for different components since we do not compute LDA among blobs of the same class.} on samples of each class separately, it is hard to control the covariances of blobs of different classes. This does not meet the homoscedasticity assumption of LDA.
In the current design, we apply LDA\cite{scikit-learn} to separate the blobs even if they do not share the same covariances. Improvement is possible by adopting heteroscedastic variants of LDA \cite{GYAMFI201744}.

\subsection{Pruning of Partitioning Hyperplanes}\label{subsec:pruning}

Stage 1 in our design gives an upper bound on the neuron and link
numbers at the first intermediate layer. To give an example, we have 4
Gaussian blobs in Example 1 while the presented MLP design has 2 LDA
systems only. It is significantly lower than the upper bound -
$6=C^4_2$.  The number of LDA systems can be reduced because some
partitioning hyperplanes are shared in splitting multiple pairs.  In
Example 1, one horizontal line partitions two Gaussian pairs and one
vertical line also partitions two Gaussian pairs. Four hyperplanes
degenerate to two. Furthermore, the 45- and 135-degree lines can be deleted
since the union of the horizontal and the vertical lines achieves the
same effect.  We would like to emphasize that the redundant design may
not affect the classification performance of the MLP. In other words,
pruning may not be essential. Yet, it may be desired to reduce the MLP model
size with little training accuracy degradation in some
applications. For this reason, we present a systematic way to reduce the
link number between $l_{in}$ and $l_1$ and the neuron number $D_1$ in $l_1$
here. 

We begin with a full design that has $M=G(G-1)/2$ LDA systems in
Stage 1. Thus, $D_1=2M$ and the number of links between $l_{in}$ and
$l_1$ is equal to $2NM$. We develop a hyperplane pruning process based
on the importance of an LDA system based on the following steps.
\begin{enumerate}
\item Delete one LDA and keep remaining LDAs the same in Stage 1.
The input space can still be split by them.
\item For a partitioned subspace enclosed by partitioning hyperplanes,
we use the majority class as the prediction for all samples in the
subspace. Compute the total number of misclassified samples in the
training data.  
\item Repeat Steps 1 and 2 for each LDA. Compare the number of
misclassified samples due to each LDA deletion. Rank the importance of
each LDA based on the impact of its deletion. An LDA is more important
if its deletion yields a higher error rate. We can delete the "least important" LDA if the resulted error rate is lower than a pre-defined threshold.
\end{enumerate}
Since there might exist correlations between multiple deleted LDA
systems, it is risky to delete multiple LDA systems simultaneously.
Thus, we delete one partitioning hyperplane (equivalently, one LDA) at a
time, run the pruning algorithm again, and evaluate the next LDA for
pruning. This process is repeated as long as the minimum resulted error rate is less than
a pre-defined threshold (and there is more than one remaining LDA).  The error threshold is a hyperparameter that
balances the network model size and the training accuracy. 

It is worth noting that it is possible that one neuron in $l_2$ covers multiple Gaussian blobs of the same class after pruning, since the separating hyperplanes between Gaussian blobs of the same class may have little impact on the classification accuracy.\footnote{In our implementation, we do not perform LDA between Gaussian blobs of the same class in Stage 1 in order to save computation.}

\subsection{Summary of FF-MLP Architecture and Link Weights}\label{subsec:FF_MLP_summary}

We can summarize the discussion in the section below. The FF-MLP
architecture and its link weights are fully specified by parameter $L$,
which is the number of partitioning hyperplanes, as follows. 
\begin{enumerate}
\item the number of the first layer neurons is $D_1=2L$,
\item the number of the second layer is $D_2 \leq 2^L$,
\item link weights in Stage 1 are determined by each individual 2-class LDA,
\item link weights in Stage 2 are either 1 or -P, and
\item link weights in Stage 3 are either 1 or 0.
\end{enumerate}

\section{Illustrative Examples}\label{sec:examples}

\begin{figure}[t]
\centering
\begin{minipage}[b]{0.49\linewidth}
\centerline{\includegraphics[width=1.0\linewidth]{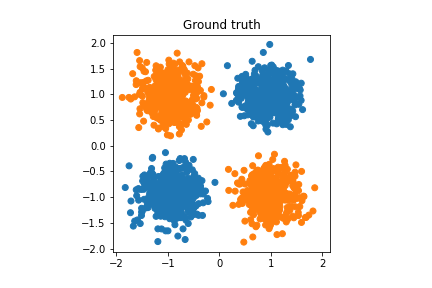}}
\centerline{(a)}
\end{minipage}
\begin{minipage}[b]{0.49\linewidth}
\centerline{\includegraphics[width=1.0\linewidth]{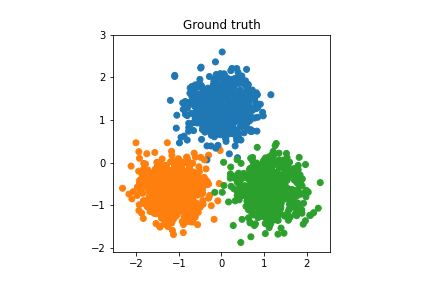}}
\centerline{(b)}
\end{minipage}
\begin{minipage}[b]{0.49\linewidth}
\centerline{\includegraphics[width=1.0\linewidth]{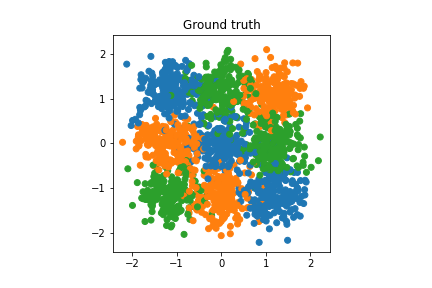}}
\centerline{(c)}
\end{minipage}
\begin{minipage}[b]{0.49\linewidth}
\centerline{\includegraphics[width=1.0\linewidth]{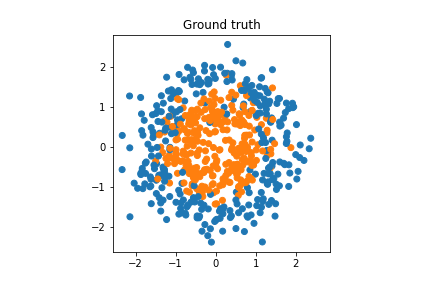}}
\centerline{(d)}
\end{minipage}
\begin{minipage}[b]{0.49\linewidth}
\centerline{\includegraphics[width=1.0\linewidth]{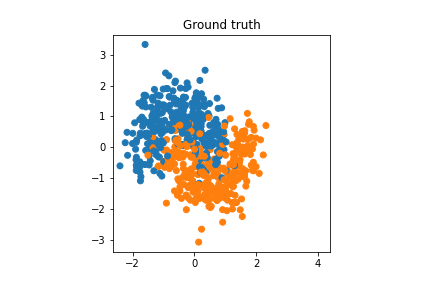}}
\centerline{(e)}
\end{minipage}
\begin{minipage}[b]{0.49\linewidth}
\centerline{\includegraphics[width=1.0\linewidth]{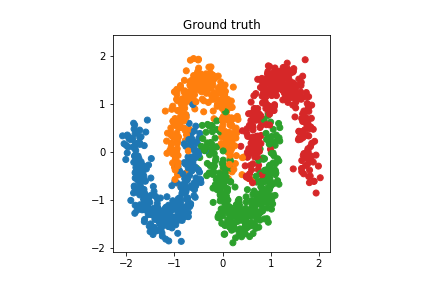}}
\centerline{(f)}
\end{minipage}
\caption{Training sample distributions of 2D examples: (a) Example 1: XOR,
(b) Example 2: 3-Gaussian-blobs, (c) Example 3: 9-Gaussian-blobs,
(d) Example 4: circle-and-ring, (e) Example 5: 2-new-moons and (f)
Example 6: 4-new-moons, where different classes are in different 
colors.}\label{fig:2d_gt}
\end{figure}

In this section, we provide more examples to illustrate the FF-MLP
design procedure. The training sample distributions of several 2D examples
are shown in Fig. \ref{fig:2d_gt}. 

\begin{figure*}[hbtp]
\centering
\begin{minipage}[b]{0.8\linewidth}
\centerline{\includegraphics[width=1.0\linewidth]{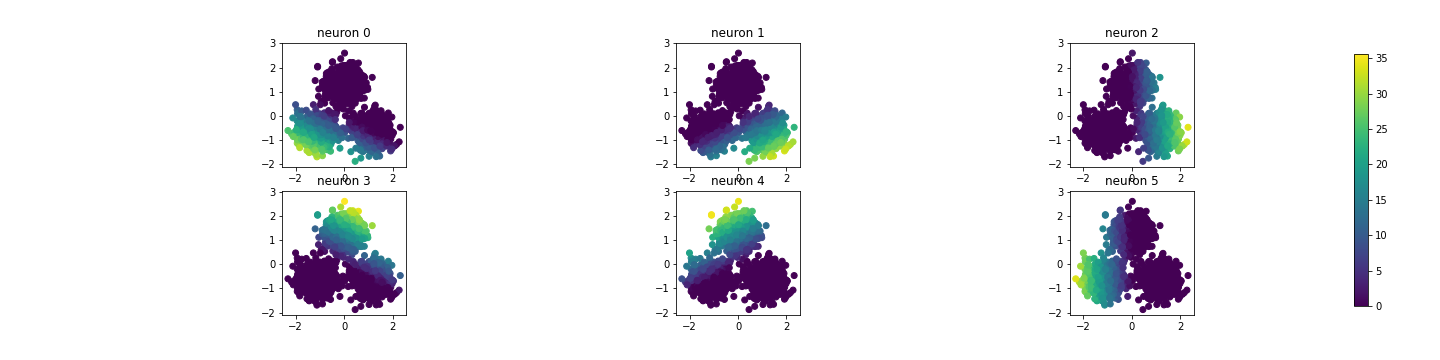}}
\centerline{(a) 1st intermediate layer}
\end{minipage}
\begin{minipage}[b]{0.8\linewidth}
\centerline{\includegraphics[width=1.0\linewidth]{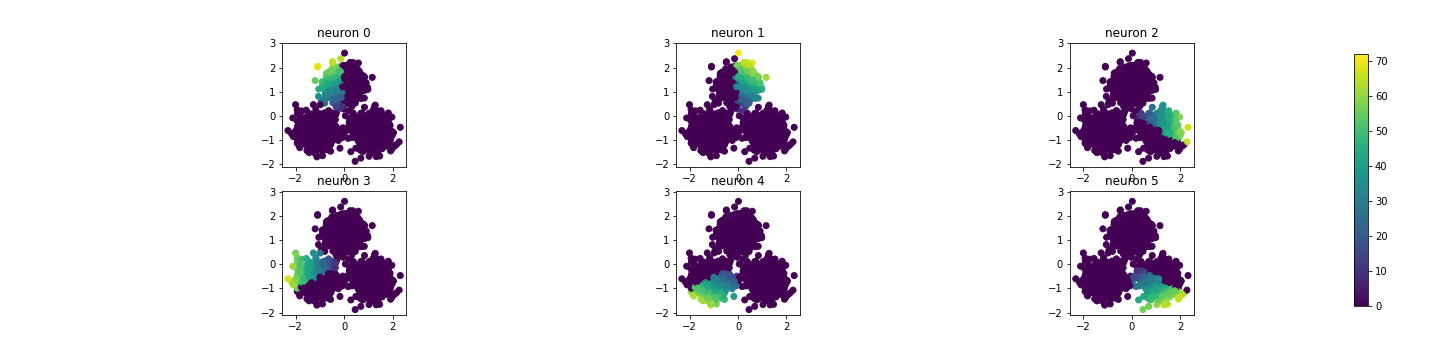}}
\centerline{(b) 2nd intermediate layer}
\end{minipage}
\caption{Response heatmaps for 3 Gaussian blobs in Example 2 in (a) layer $l_1$
and (b) layer $l_2$.}\label{fig:3gaussian_heatmaps}
\end{figure*}

\noindent
{\bf Example 2 (3-Gaussian-blobs)}. There are three Gaussian-distributed
classes in blue, orange and green as shown in Fig. \ref{fig:2d_gt}(b). The three Gaussian blobs have identical covariance matrices.
We use three lines and $D_1=6$ neurons in layer
$l_1$ to separate them in Stage 1.  Fig.
\ref{fig:3gaussian_heatmaps}(a) shows neuron responses in layer $l_1$.
We see three neuron pairs: 0 and 3, 1 and 4, and 2 and 5.  In Stage 2,
we would like to isolate each Gaussian blob in a certain subspace.
However, due to the shape of activated regions in Fig.
\ref{fig:3gaussian_heatmaps}(b), we need two neurons to preserve one
Gaussian blob in layer $l_2$. For example, neurons 0 and 1 can preserve
the Gaussian blob in the top as shown in Fig.
\ref{fig:3gaussian_heatmaps}(b). 

\begin{figure}[htpb]
\centering
\begin{minipage}[b]{0.49\linewidth}
\centerline{\includegraphics[width=1.0\linewidth]{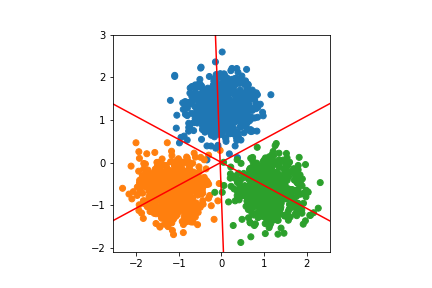}}
\centerline{(a) Example 2}
\end{minipage}
\begin{minipage}[b]{0.49\linewidth}
\centerline{\includegraphics[width=1.0\linewidth]{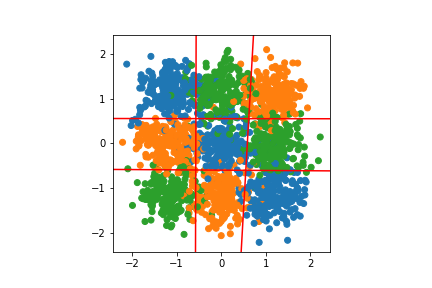}}
\centerline{(b) Example 3}
\end{minipage}
\caption{Partitioning lines for (a) Example 2 and (b) Example 3.}\label{fig:hyperplanes}
\end{figure}

If three partitioning lines $h_1$, $h_2$, $h_3$ intersect at nearly the same
point as illustrated in Fig. \ref{fig:hyperplanes}(a), we have 6 nonempty regions
instead of 7. Our FF-MLP design has $D_{in}=2$, $D_1=6$, $D_2=6$ and
$D_{out}=3$. The training accuracy and the testing accuracy of the
designed FF-MLP are 99.67\% and 99.33\%, respectively.  This shows that
the MLP splits the training data almost perfectly and fits the
underlying data distribution very well. 

\begin{figure*}[htpb]
\centering
\begin{minipage}[b]{0.8\linewidth}
\centerline{\includegraphics[width=1.0\linewidth]{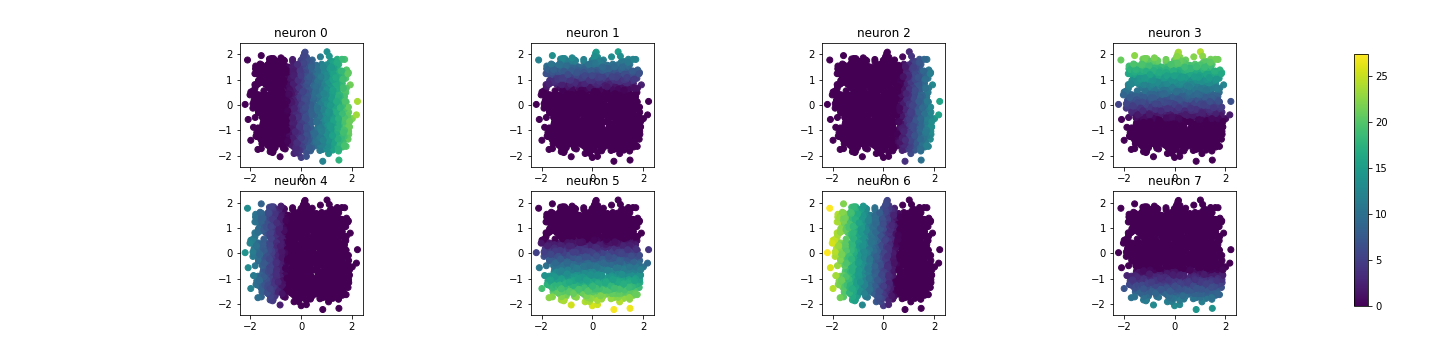}}
\centerline{(a)}
\end{minipage}
\begin{minipage}[b]{0.8\linewidth}
\centerline{\includegraphics[width=1.0\linewidth]{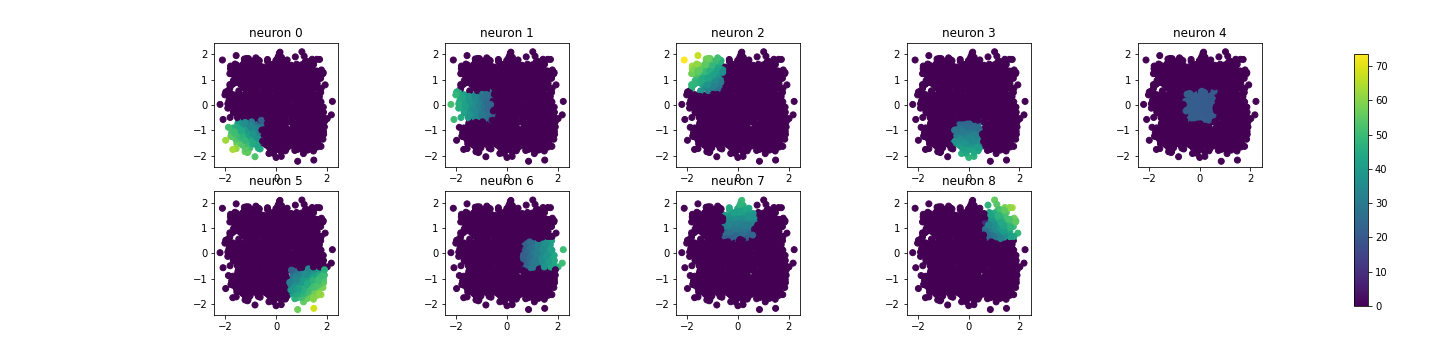}}
\centerline{(b)}
\end{minipage}
\caption{Neuron responses of the 9-Gaussian-blobs in Example 3 in 
(a) layer $l_1$ and (b) layer $l_2$.}\label{fig:9gaussian_4lines_heatmaps}
\end{figure*}

\begin{figure}[htpb]
\centering
\begin{minipage}[b]{0.49\linewidth}
\centerline{\includegraphics[width=1.0\linewidth]{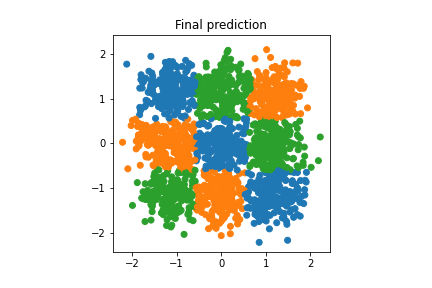}}
\centerline{(a) Th=0.3}
\end{minipage}
\begin{minipage}[b]{0.49\linewidth}
\centerline{\includegraphics[width=1.0\linewidth]{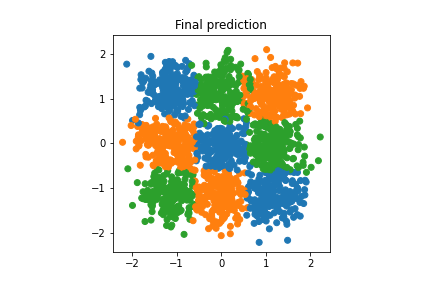}}
\centerline{(b) Th=0.1}
\end{minipage}
\caption{Comparison of classification results of FF-MLP for Example 3 
with two different error thresholds: (a) Th=0.3 and (b) Th=0.1.} 
\label{fig:9gaussian_gtpredcomp_ff}
\end{figure}

\noindent
{\bf Example 3 (9-Gaussian-blobs)}. It contains 9 Gaussian blobs of the
same covariance matrices aligned in 3 rows and 3 columns as shown in
Fig.  \ref{fig:2d_gt}(c). In Stage 1, we need $C^9_2= 36$ separating
lines\footnote{In implementation, we only generate 27 lines to separate blobs of different classes.} at most.  We run the pruning algorithm with the error threshold
equal to 0.3 and reduce the separating lines to 4, each of which
partition adjacent rows and columns as expected.  Then, there are
$D_1=8$ neurons in layer $l_1$.  The neuron responses in layer $l_1$ are
shown in Fig.  \ref{fig:9gaussian_4lines_heatmaps}(a). They are grouped
in four pairs: 0 and 4, 1 and 5, 2 and 6, and 3 and 7.  In Stage 2, we
form 9 regions, each of which contains one Gaussian blob as shown in
Fig.  \ref{fig:9gaussian_4lines_heatmaps}(b).  As shown in Fig.
\ref{fig:hyperplanes}, we have two pairs of nearly parallel partitioning lines.
Only 9 nonempty regions are formed in the finite range. The training and
testing accuracy are 88.11\% and 88.83\% respectively.  The error
threshold affects the number of partitioning lines.  When we set the
error threshold to 0.1, we have 27 partitioning lines and $54$ neurons
in layer $l_1$. By doing so, we preserve all needed boundaries between
blobs of different classes. The training and testing accuracy are
89.11\% and 88.58\%, respectively. The performance difference is very
small in this case.

\begin{figure}[htpb]
\centering
\begin{minipage}[b]{0.49\linewidth}
\centerline{\includegraphics[width=1.0\linewidth]{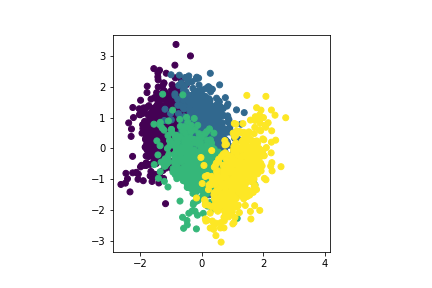}}
\centerline{(a)}
\end{minipage}
\begin{minipage}[b]{0.49\linewidth}
\centerline{\includegraphics[width=1.0\linewidth]{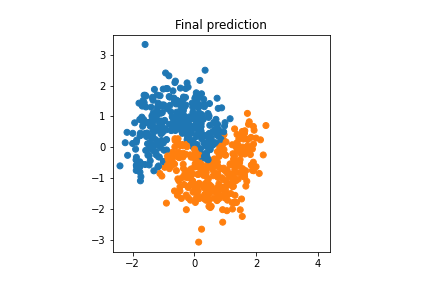}}
\centerline{(b)}
\end{minipage}
\caption{Visualization of the 2-new-moons example: (a) the generated random samples from the fitted GMMs with 2 components per class and (b) the classification result of FF-MLP with error threshold equal to 0.1.}\label{fig:two_new_moon}
\end{figure}

\begin{figure}[htpb]
\centering
\includegraphics[width=0.49\linewidth]{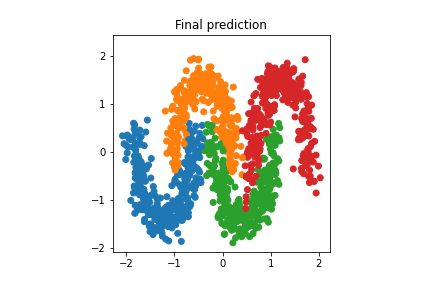}
\caption{Classification results of FF-MLP for Example 6, where each new moon
is approximated by 3 Gaussian components.}\label{fig:4moon_gtpredcomp}
\end{figure}

\begin{figure}[thpb]
\centering
\begin{minipage}[b]{0.49\linewidth}
\centerline{\includegraphics[width=1.0\linewidth]{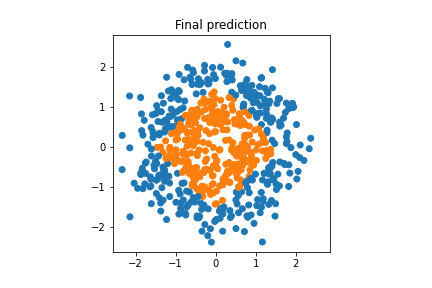}}
\centerline{(a) FF-MLP, 4 Components}
\end{minipage}
\begin{minipage}[b]{0.49\linewidth}
\centerline{\includegraphics[width=1.0\linewidth]{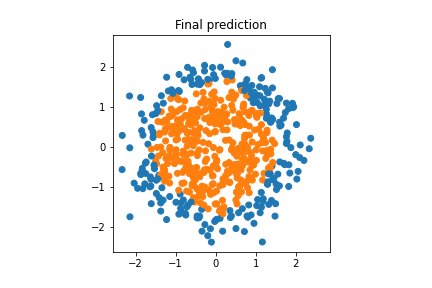}}
\centerline{(b) FF-MLP, 16 Components}
\end{minipage}
\begin{minipage}[b]{0.49\linewidth}
\centerline{\includegraphics[width=1.0\linewidth]{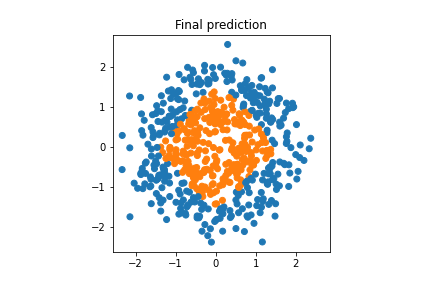}}
\centerline{(c) BP-MLP, 4 Components}
\end{minipage}
\begin{minipage}[b]{0.49\linewidth}
\centerline{\includegraphics[width=1.0\linewidth]{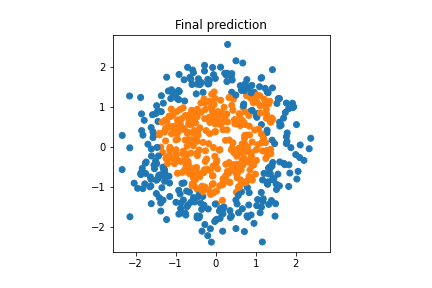}}
\centerline{(d) BP-MLP, 16 Components}
\end{minipage}
\caption{Classification results for the circle-and-ring example with (a)
FF-MLP, 4 components, (b) FF-MLP, 16 components, (c) BP-MLP with FF-MLP initialization, 4
components, and (d) BP-MLP with FF-MLP initialization, 16 components.}\label{fig:circle_MLP}
\end{figure}

\noindent
{\bf Example 4 (Circle-and-Ring).} It contains an inner circle as one
class and an outer ring as the other class as shown in Fig.
\ref{fig:2d_gt}(d)\cite{scikit-learn}. To apply our MLP design, we use one Gaussian blob to
model the inner circle and approximate the outer ring with 4 and 16
Gaussian components, respectively. For the case of 4 Gaussian
components, blobs of different classes can be separated by 4
partitioning lines. By using a larger number of blobs, we may obtain a
better approximation to the original data.  The corresponding
classification results are shown in Figs.  \ref{fig:circle_MLP}(a)
and (b). We see that the decision boundary of 16 Gaussian components 
is smoother than that of 4 Gaussian components. 

\noindent 
{\bf Example 5 (2-New-Moons).} It contains two interleaving new moons
as shown in Fig. \ref{fig:2d_gt}(e)\cite{scikit-learn}. Each new moon corresponds to one
class.  We use 2 Gaussian components for each class and show the generated samples from the fitted GMMs in Fig.  \ref{fig:two_new_moon}(a), which appears
to be a good approximation to the original data visually.  By applying our design to the
Gaussian blobs, we obtain the classification result as shown in Fig.
\ref{fig:two_new_moon}(b), which is very similar to the ground truth
(see Table \ref{tab:acc_comp}).

\noindent
{\bf Example 6 (4-New-Moons).} It contains four interleaving new
moons as shown in Fig. \ref{fig:2d_gt}(f)\cite{scikit-learn}, where each moon is a class.
We set the number of blobs to 3 for each moon and the error
threshold to 0.05. There are 9 partitioning lines and 18 neurons in layer $l_1$, which in turn yields 28 region neurons in layer $l_2$.  The classification results are shown in Fig. \ref{fig:4moon_gtpredcomp}.  We can see that the predictions are similar to the ground truth and fit the underlying distribution quite well.  The training accuracy is
95.75\% and the testing accuracy is 95.38\%. 

\section{Observations on BP-MLP Behavior}\label{sec:insights}

Even when FF-MLP and BP-MLP adopt the same MLP architecture designed by our
proposed method, BP-MLP has two differences from FF-MLP: 1)
backpropagation (BP) optimization of the cross-entropy loss function and
2) network initialization. We report observations on the effects of BP
and different initializations in Secs.  \ref{subsec:impact_bp} and
\ref{sec:bp_rand}, respectively. 

\begin{figure}[htpb]
\centering
\begin{minipage}[b]{1.0\linewidth}
\centerline{\includegraphics[width=0.6\linewidth]{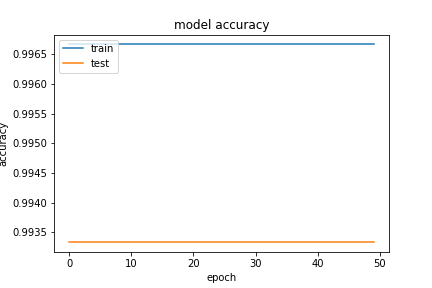}}
\centerline{(a) 3-Gaussian-blobs}
\end{minipage}
\begin{minipage}[b]{1.0\linewidth}
\centerline{\includegraphics[width=0.6\linewidth]{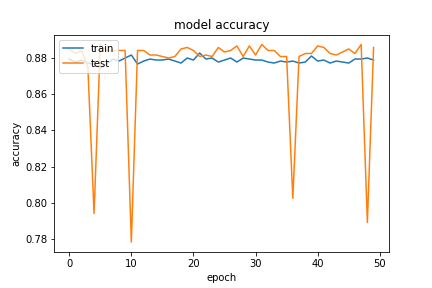}}
\centerline{(b) 9-Gaussian-blobs, Th=0.3}
\end{minipage}
\begin{minipage}[b]{1.0\linewidth}
\centerline{\includegraphics[width=0.6\linewidth]{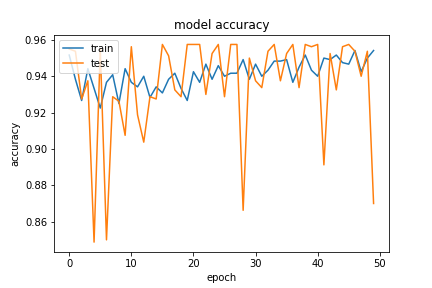}}
\centerline{(c) 4-new-moons}
\end{minipage}
\begin{minipage}[b]{1.0\linewidth}
\centerline{\includegraphics[width=0.6\linewidth]{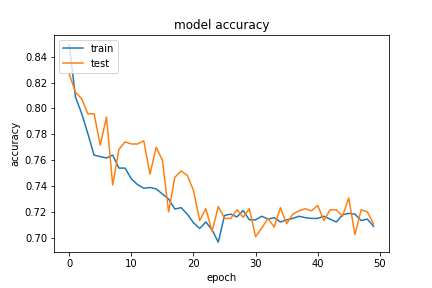}}
\centerline{(d) 9-Gaussian-blobs, Th=0.1}
\end{minipage}
\begin{minipage}[b]{1.0\linewidth}
\centerline{\includegraphics[width=0.6\linewidth]{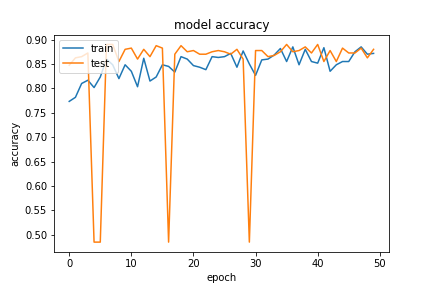}}
\centerline{(e) circle-and-ring}
\end{minipage}
\caption{Training and testing accuracy curves of BP-MLP as functions of
the epoch number for (a) 3-Gaussian-blobs, (b) 9-Gaussian-blobs (Th=0.3), (c)
4-new-moons, (d) 9-Gaussian-blobs (Th=0.1) and (e) circle-and-ring, where the network is initialized by the proposed FF-MLP.}
\label{fig:bp_acc}
\end{figure}

\subsection{Effect of Backpropagation (BP)}\label{subsec:impact_bp} 

We initialize a BP-MLP with weights and biases of the FF-MLP design and
conduct BP using the gradient descent (SGD) optimizer with
0.01 learning rate and zero momentum. We observe four representative
cases, and show the training and testing accuracy curves as a function
of the epoch number in Figs. \ref{fig:bp_acc}(a)-(e). 
\begin{itemize}
\item BP has very little effect. \\
One such example is the 3-Gaussian-blobs case. Both training and
testing curves remain at the same level as shown in Fig. \ref{fig:bp_acc}(a).
\item BP has little effect on training but a negative effect on testing. \\
The training and testing accuracy curves for the 9-Gaussian-blobs case are
plotted as a function of the epoch number in Fig. \ref{fig:bp_acc}(b). The network has 8 neurons in $l_1$ and 9 neurons in $l_2$, which is the same network architecture as in Fig.
\ref{fig:9gaussian_gtpredcomp_ff}(a).
Although the training accuracy remains at the same level, the testing
accuracy fluctuates with several drastic drops. This behavior is
difficult to explain, indicating the interpretability challenge of
BP-MLP. 
\item BP has negative effects on both training and testing. \\
The training and testing accuracy curves for 4-new-moons case are
plotted as a function of the epoch number in Fig. \ref{fig:bp_acc}(c).
Both training and testing accuracy fluctuate and several drastic drops are observed for the testing accuracy. As shown in Table \ref{tab:acc_comp}, the final training and testing accuracy are lower compared to the FF-MLP results. The predictions for the training samples are shown in Fig. \ref{fig:4moon_BPMLP_predcomp}(a), which is worse than the ones in Fig. \ref{fig:4moon_gtpredcomp}. 
Another example is the 9-Gaussian-blobs case with error threshold equal to 1. The training and testing accuracy decrease during training. The final training and testing accuracy are lower than FF-MLP as shown in Table \ref{tab:acc_comp}. 
\item BP has positive effect on both training and testing. \\
For the circle-and-ring example when the outer ring is approximated with 16 components, the predictions for the training samples after BP are shown in Fig. \ref{fig:circle_MLP}(d). 
We can see the improvement in the training and testing accuracy in Table \ref{tab:acc_comp}. However, similar to the 9-Gaussian-blobs and 4-new-moons cases, we also observe the drastic drops of the testing accuracy during training.
\end{itemize}

\subsection{Effect of Initializations}\label{sec:bp_rand}

\begin{figure}[htpb]
\centering
\begin{minipage}[b]{0.49\linewidth}
\centerline{\includegraphics[width=1.0\linewidth]{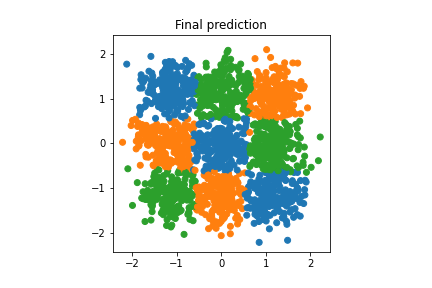}}
\centerline{(a) FF-MLP initialization}
\end{minipage}
\begin{minipage}[b]{0.49\linewidth}
\centerline{\includegraphics[width=1.0\linewidth]{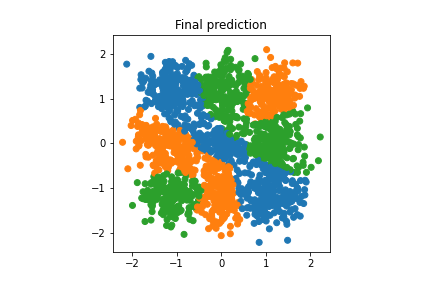}}
\centerline{(b) random initialization}
\end{minipage}
\caption{Comparison of BP-MLP classification results for
9-Gaussian-blobs with (a) FF-MLP initialization and (b) random
initialization.}\label{fig:9gaussian_gtpredcomp_bp}
\end{figure}

\begin{figure}[htpb]
\centering
\begin{minipage}[b]{0.49\linewidth}
\centerline{\includegraphics[width=1.0\linewidth]{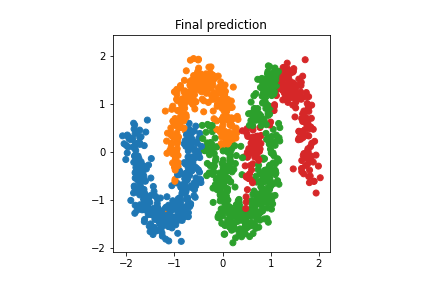}}
\centerline{(a) FF-MLP initialization}
\end{minipage}
\begin{minipage}[b]{0.49\linewidth}
\centerline{\includegraphics[width=1.0\linewidth]{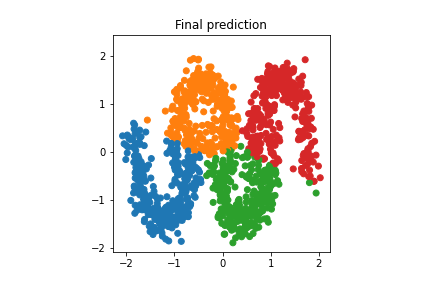}}
\centerline{(b) random initialization}
\end{minipage}
\caption{Comparison of BP-MLP classification results for 4-new-moons
with (a) FF-MLP initialization and (b) random
initialization.}\label{fig:4moon_BPMLP_predcomp}
\end{figure}

We compare different initialization schemes for BP-MLP. One is to
use FF-MLP and the other is to use the random initialization. 
We have the following observations.
\begin{itemize}
\item Either initialization works. \\
For some datasets such as the 3-Gaussian-blob data, their final classification 
performance is similar using either random initialization or FF-MLP initialization. 
\item In favor of the FF-MLP initialization. \\
We compare the BP-MLP classification results for 9-Gaussian-blobs with
FF-MLP and random initializations in Fig.
\ref{fig:9gaussian_gtpredcomp_bp}. The network has 8 neurons in $l_1$ and 9 neurons in $l_2$, which is the same network architecture as in Fig.
\ref{fig:9gaussian_gtpredcomp_ff}(a).
The advantage of the FF-MLP
initialization is well preserved by BP-MLP. In contrast, with the random
initialization, BP tends to find smooth boundaries to split data,
which does not fit the underlying source data distribution in this case.
\item  Both initializations fail. \\
We compare the BP-MLP classification results for 4-new-moons with FF-MLP
and random initializations in Fig. \ref{fig:4moon_BPMLP_predcomp}.   The result with the random initialization fails to capture the concave moon shape.
\end{itemize}
Generally speaking, BP-MLP with random initialization tends to
over-simplify the boundaries and data distribution as observed in both
9-Gaussian-blobs and 4-new-moons. 

\section{Experiments}\label{sec:experiments}

\subsection{Classification Accuracy for 2D Samples}\label{subsec:2D_exp}

We compare training and testing classification performance among FF-MLP,
BP-MLP with FF-MLP initialization, and BP-MLP with random initialization
for Examples 1-6 in the last section in Table \ref{tab:acc_comp}. The
integers in parentheses in the first row for BP-MLP are the epoch
numbers.  In the first column, the numbers in parentheses for the
9-Gaussian-blobs are error thresholds, the numbers in parentheses for
the circle-and-ring are Gaussian component numbers for outer-ring's
approximations.  For BP-MLP with random initialization, we report
means and standard deviations of classification accuracy over 5 runs. We used the Xavier uniform initializer \cite{glorot2010understanding} for random initialization. We trained the network for two different epoch numbers; namely, 15 epochs and 50 epochs in different runs. 


\begin{table*}[tbhp]
\centering
\begin{tabular}{c | c c | c c | c c | c c}\hline
\multirow{2}{*}{Dataset} & \multicolumn{2}{c|}{FF-MLP} & \multicolumn{2}{c|}{BP-MLP with FF-MLP init.(50)} & \multicolumn{2}{c|}{BP-MLP with random init. (50)}  & \multicolumn{2}{c}{BP-MLP with random init. (15)}\\
\cline{2-9}
 & train & test & train & test & train & test & train & test\\ \hline
XOR & \textbf{100.00} & \textbf{99.83} & \textbf{100.00} & \textbf{99.83} & 99.83 $\pm$ 0.16 & 99.42 $\pm$ 0.24 & 93.20 $\pm$ 11.05 & 92.90 $\pm$ 11.06\\
3-Gaussian-blobs & 99.67 & 99.33 & 99.67 & 99.33 & \textbf{99.68} $\pm$ 0.06 & \textbf{99.38} $\pm$ 0.05 & 99.48 $\pm$ 0.30 & 99.17 $\pm$ 0.48\\
9-Gaussian-blobs (0.1) & \textbf{89.11} & \textbf{88.58} & 70.89 & 71.08 & 84.68 $\pm$ 0.19 & 85.75 $\pm$ 0.24 & 78.71 $\pm$ 2.46 & 78.33 $\pm$ 3.14\\
9-Gaussian-blobs (0.3) & \textbf{88.11} & \textbf{88.83} & 88.06 & 88.58 & 81.62 $\pm$ 6.14 & 81.35 $\pm$ 7.29 & 61.71 $\pm$ 9.40 & 61.12 $\pm$ 8.87\\
circle-and-ring (4) & 88.83 & \textbf{87.25} & \textbf{89.00} & 86.50 & 81.93 $\pm$ 7.22 & 82.80 $\pm$ 5.27 & 70.57 $\pm$ 13.42 & 71.25 $\pm$ 11.27\\
circle-and-ring (16) & 83.17 & 80.50 & 85.67 & \textbf{88.00} & \textbf{86.20} $\pm$ 1.41 & 85.05 $\pm$ 1.85 & 66.20 $\pm$ 9.33 & 65.30 $\pm$ 11.05\\
2-new-moons & \textbf{88.17} & \textbf{91.25} & \textbf{88.17} & \textbf{91.25} & 83.97 $\pm$ 1.24 & 87.60 $\pm$ 0.52 & 82.10 $\pm$ 1.15 & 86.60 $\pm$ 0.58\\
4-new-moons & \textbf{95.75} & \textbf{95.38} & 87.50 & 87.00 & 86.90 $\pm$ 0.25 & 84.00 $\pm$ 0.33 & 85.00 $\pm$ 0.98 & 82.37 $\pm$ 0.76\\
\hline
\end{tabular}
\caption{Comparison of training and testing classification performance
between FF-MLP, BP-MLP with FF-MLP initialization and BP-MLP with random
initialization. The best (mean) training and testing accuracy are
highlighted in bold.}\label{tab:acc_comp}
\end{table*}

Let us focus on test accuracy. FF-MLP has better test performance with the 
following settings:
\begin{itemize}
\item XOR (99.83\%)
\item 9-Gaussian-blobs with error threshold 0.1 or 0.3 (88.58\% and 88.83\% respectively);
\item circle-and-ring with 4 Gaussian components for the outer ring (87.25\%);
\item 2-new-moons (91.25\%)
\item 4-new-moons (95.38\%).
\end{itemize}
Under these settings, FF-MLP outperforms BP-MLP with FF-MLP initialization or random initialization. 

\subsection{Classification Accuracy for Higher-Dimensional Samples}\label{subsec:HD_exp}

Besides 2D samples, we test FF-MLP for four higher-dimensional datasets.  
The datasets are described below. 
\begin{itemize}
\item {\bf Iris Dataset.} The Iris plants dataset \cite{scikit-learn,fisher1936use} is
a classification dataset with 3 different classes and 150 samples in total. The input dimension is 4.  
\item {\bf Wine Dataset.} The Wine recognition dataset \cite{scikit-learn,Dua:2019} has 3
classes with 59, 71, and 48 samples respectively.  The input dimension is 13.  
\item {\bf B.C.W. Dataset.} The breast cancer wisconsin (B.C.W) dataset
\cite{scikit-learn,Dua:2019} is a binary classification dataset. It has 569 samples
in total. The input dimension is 30. 
\item {\bf Pima Indians Diabetes Dataset.} The Pima Indians diabetes dataset\footnote{We used the data from https://www.kaggle.com/uciml/pima-indians-diabetes-database for our experiments.}
\cite{smith1988using} is for diabetes prediction. It is a binary
classification dataset with 768 8-dimensional entries. In our experiments, we removed the samples with the physically impossible value zero for glucose, diastolic blood pressure, triceps skin fold thickness, insulin, or BMI. We used only the remaining 392 samples.
\end{itemize}

We report the training and testing accuracy results of FF-MLP, BP-MLP
with random initialization and trained with 15 and 50 epochs in Table
\ref{table:highD}. For BP-MLP, the means of classification
accuracy and the standard deviations over 5 runs are reported. 

\begin{table*}[thbp]
\centering
\begin{tabular}{c | c | c | c | c | c | c | c | c | c | c} 
\hline
\multirow{3}{*}{Dataset} & \multirow{3}{*}{$D_{in}$} & \multirow{3}{*}{$D_{out}$} & \multirow{3}{*}{$D_1$} & \multirow{3}{*}{$D_2$} &
\multicolumn{6}{c}{Accuracy} \\
\cline{6-11}
&&&&&\multicolumn{2}{c|}{FF-MLP} & \multicolumn{2}{c|}{BP-MLP/random init. (50)} & \multicolumn{2}{c}{BP-MLP/random init. (15)}\\
\cline{6-11}
 &  & & & & train & test & train & test & train & test \\ \hline
Iris & 4 & 3 & 4 & 3 & \textbf{96.67} & \textbf{98.33} & 65.33 $\pm$ 23.82 & 64.67 $\pm$ 27.09 & 47.11 $\pm$ 27.08 & 48.33 $\pm$ 29.98 \\
Wine & 13 & 3 & 6 & 6 & \textbf{97.17} & \textbf{94.44} & 85.66 $\pm$ 4.08 & 79.72 $\pm$ 9.45 & 64.34 $\pm$ 7.29 & 61.39 $\pm$ 8.53 \\
B.C.W & 30 & 2 & 2 & 2 & \textbf{96.77} & 94.30 & 95.89 $\pm$ 0.85 & \textbf{97.02} $\pm$ 0.57 & 89.79 $\pm$ 2.41 & 91.49 $\pm$ 1.19 \\
Pima & 8 & 2 & 18 & 88 & \textbf{91.06} & 73.89 & 80.34 $\pm$ 1.74 & \textbf{75.54} $\pm$ 0.73 & 77.02 $\pm$ 2.89 & 73.76 $\pm$ 1.45 \\[1ex]
\hline
\end{tabular}
\caption{Training and testing accuracy results of FF-MLP and BP-MLP with
random initialization for four higher-dimensional datasets. The best
(mean) training and testing accuracy are highlighted in bold.}\label{table:highD}
\end{table*}

For the Iris dataset, we set the number of Gaussian components to 2 in
each class. The error threshold is set to 0.05.  There are 2
partitioning hyperplanes and, thus, 4 neurons in layer $l_1$.  The
testing accuracy for FF-MLP is 98.33\%.  For BP-MLP with random
initialization trained for 50 epochs, the mean test accuracy is
64.67\%. In this case, it seems that the network may have been too small for BP-MLP to arrive at a good solution.

For the wine dataset, we set the number of Gaussian components to 2 in
each class. The error threshold is set to 0.05. There are 3 partitioning
hyperplanes and thus 6 neurons in layer $l_1$.  The testing accuracy
for FF-MLP is 94.44\%. Under the same architecture, BP-MLP with random
initialization and 50 epochs gives the mean test accuracy of 79.72\%.
FF-MLP outperforms BP-MLP. 

For the B.C.W. dataset, we set the number of Gaussian components to 2
per class. The error threshold is set to 0.05. There are 1 partitioning
hyperplanes and thus 2 neurons in layer $l_1$.  The testing accuracy for
FF-MLP is 94.30\%. The mean testing accuracy for BP-MLP is 97.02\%.
BP-MLP outperforms FF-MLP on this dataset. 

Finally, for the Pima Indians dataset, we set the number of
Gaussian components to 4 per class. The error threshold is set to 0.1.
There are 9 partitioning hyperplanes and thus 18 neurons in layer
$l_1$. The testing accuracy for FF-MLP is 73.89\% while that for BP-MLP
is 75.54\% on average. BP-MLP outperforms FF-MLP in terms of testing accuracy. 

\subsection{Computational Complexity}\label{subsec:complexity}

We compare the computational complexity of FF-MLP and BP-MLP in Table
\ref{table:running_time}, Tesla V100-SXM2-16GB GPU was used in the
experiments. For FF-MLP, we show the time of each step of our design and
its sum in the total column.  The steps include: 1) preprocessing with
GMM approximation (stage 0), 2) fitting partitioning hyperplanes and
redundant hyperplane deletion (stage 1), 3) Gaussian blob isolation
(stage 2), and 4) assigning each isolated region to its corresponding
output class (stage 3). For comparison, we show the training time of
BP-MLP in 15 epochs and 50 epochs (in two separate runs) under the same
network with random initialization. 

\begin{table*}[thbp]
\centering
\begin{tabular}{c | c c c c c | c c} \hline
\multirow{2}{*}{Dataset} & \multirow{2}{*}{GMM} & Boundary & Region & Classes & \multirow{2}{*}{\textbf{Total}} & \multirow{2}{*}{BP (15)} & \multirow{2}{*}{BP (50)}\\
& & construction & representation & assignment & & &\\  \hline  \hline
XOR & 0.00000 & 0.01756 & 0.00093 & 0.00007 & \textbf{0.01855} & 2.88595 $\pm$ 0.06279 & 8.50156 $\pm$ 0.14128\\
3-Gaussian-blobs & 0.00000 & 0.01119 & 0.00126 & 0.00008 & \textbf{0.01253} & 2.78903 $\pm$ 0.07796 & 8.26536 $\pm$ 0.17778\\
9-Gaussian-blobs (0.1) & 0.00000 & 0.22982 & 0.00698 & 0.00066 & \textbf{0.23746} & 2.77764 $\pm$ 0.14215 & 8.34885 $\pm$ 0.28903\\
9-Gaussian-blobs (0.3) & 0.00000 & 2.11159 & 0.00156 & 0.00010 & \textbf{2.11325} & 2.79140 $\pm$ 0.06179 & 8.51242 $\pm$ 0.24676\\
circle-and-ring (4) & 0.02012 & 0.01202 & 0.00056 & 0.00006 & \textbf{0.03277} & 1.50861 $\pm$ 0.14825 & 3.79068 $\pm$ 0.28088\\
circle-and-ring (16) & 0.04232 & 0.05182 & 0.00205 & 0.00020 & \textbf{0.09640} & 1.43951 $\pm$ 0.15573 & 3.80061 $\pm$ 0.13775\\
2-new-moons & 0.01835 & 0.01111 & 0.00053 & 0.00006 & \textbf{0.03006} & 1.44454 $\pm$ 0.06723 & 3.64791 $\pm$ 0.08565\\
4-new-moons & 0.03712 & 11.17161 & 0.00206 & 0.00021 & 11.21100 & \textbf{1.98338} $\pm$ 0.04357 & 5.71387 $\pm$ 0.14150\\
Iris & 0.02112 & 0.02632 & 0.00011 & 0.00002 & \textbf{0.04757} & 0.73724 $\pm$ 0.01419 & 1.60543 $\pm$ 0.14658\\
Wine & 0.01238 & 0.03551 & 0.00015 & 0.00003 & \textbf{0.04807} & 0.81173 $\pm$ 0.01280 & 1.72276 $\pm$ 0.07268\\
B.C.W & 0.01701 & 0.03375 & 0.00026 & 0.00003 & \textbf{0.05106} & 1.08800 $\pm$ 0.05579 & 2.73232 $\pm$ 0.12023\\
Pima & 0.03365 & 0.16127 & 0.00074 & 0.00039 & \textbf{0.19604} & 0.96707 $\pm$ 0.03306 & 2.32731 $\pm$ 0.10882\\
\hline
\end{tabular}
\caption{Comparison of computation time in seconds of FF-MLP (left) and
BP-MLP (right) with 15 and 50 epochs.  The mean and standard deviation
of computation time in 5 runs are reported for BP-MLP.  The shortest
(mean) running time is highlighted in bold.}\label{table:running_time}
\end{table*}

As shown in Table \ref{table:running_time} FF-MLP takes 1 second or less
in the network architecture and weight design. The only exceptions are
4-new-moons and 9-Gaussian-blobs with an error threshold of 0.3.  Most computation time is spent on boundary construction, which includes hyperplane pruning. To determine the hyperplane to prune, we need to repeat the temporary hyperplane deletion and error rate computation for each hyperplane. This is a very time-consuming process.
If we do not prune any hyperplane, the total computation time 
can be significantly shorter at the cost of a larger network size.  
Generally speaking, for non-Gaussian datasets, GMM approximation and fitting partitioning hyperplanes and redundant hyperplane deletion take
most execution time among all steps. For datasets consisting of Gaussian blobs, no GMM
approximation is needed. For BP-MLP, the time 
reported in Table \ref{table:running_time} does not include network
design time but only the training time with BP.  It is apparent from the
table that the running time in the FF-MLP design is generally shorter than
that in the BP-MLP design. 

\section{Comments on Related Work}\label{sec:comments}

We would like to comment on prior work that has some connection with our
current research.  

\subsection{BP-MLP Network Design}\label{subsec:BP-MLP_design}

Design of the BP-MLP network architectures has been studied by quite a
few researchers in the last three decades. It is worthwhile to review
efforts in this area as a contrast of our FF-MLP design.  Two approaches in the design of BP-MLPs will be examined in this and the next subsections: 1) viewing architecture design as an optimization problem and 2) adding neurons and/or layers as needed in the training process. 

\subsubsection{Architecture Design as an Optimization Problem}\label{subsubsec:search}

One straightforward idea to design the BP-MLP architecture is to try different sizes and select the one that gives the best performance. A good search algorithm is needed to reduce the trial number.
For example, Doukim {\em et al.} \cite{doukim2010finding} used a coarse-to-fine two-step search algorithm to determine the number of neurons in a hidden layer. First, a coarse
search was performed to refine the search space. Then, a fine-scale sequential search was conducted near the optimum value obtained in the first step.  
Some methods were also proposed for both structure and weight optimization.
Ludermir {\em et al.} \cite{ludermir2006optimization} used a tabu search \cite{glover1986future} and simulated annealing \cite{kirkpatrick1983optimization} based method to find the optimal architecture and parameters at the same time.  
Modified bat algorithm \cite{yang2010new} was adopted in \cite{jaddi2015optimization} to optimize the network parameters and architecture. 
The MLP network structure and its weights were optimized in \cite{zanchettin2011hybrid} based on  backpropagation, tabu search, heuristic
simulated annealing, and genetic algorithms \cite{10.5555/534133}. 

\subsubsection{Constructive Neural Network Learning}\label{subsubsec:constructive}

In constructive design, neurons 
can be added to the network as needed to achieve better performance. 
Examples include: \cite{parekh1997constructive, mezard1989learning, frean1990upstart}. 
To construct networks, an iterative weight update approach \cite{parekh2000constructive} can be adopted \cite{mezard1989learning, frean1990upstart,
stephen1990perceptron, mascioli1995constructive}. 
Neurons were sequentially added to separate multiple groups in one class
from the other class in \cite{marchand1990convergence}. Samples of one
class were separated from all samples of another class with a newly
added neuron at each iteration. Newly separated samples are excluded
from consideration in search of the next neuron. The input space is
split into pure regions that contain only samples of the same class
while the output of a hidden layer is proved to be linearly separable.

Another idea to construct networks is to leverage geometric properties \cite{parekh2000constructive},
e.g., \cite{yang1999distal, alpaydin1994gal, marchand1989learning,
bose1993neural, bennett1990neural}. Spherical threshold neurons were
used in \cite{yang1999distal}. Each neuron is activated if 
\begin{equation}
\theta_{low} \leq d(\vct{W}, \vct{X}) \leq \theta_{high}, 
\end{equation}
where $\vct{W}$, $\vct{X}$, $\theta_{low}$ and $\theta_{high}$ are the
weight vector, the input vector, lower and higher thresholds
respectively, and $d(\cdot)$ represents the distance function.  That is,
only samples between two spheres of radii $\theta_{low}$ and
$\theta_{high}$ centered at the $\vct{W}$ are activated.  At each
iteration, a newly added neuron excludes the largest number of samples
of the same class in the activated region from the working set.  The
hidden representations are linearly separable by assigning weights to
the last layer properly. This method can determine the thresholds and
weights without backpropagation. 

\subsection{Relationship with Work on Interpretation of Neurons as
Partitioning Hyperplanes}\label{subsec:partitioning}

The interpretation of neurons as partitioning hyperplanes was done by some researchers before.  As described in Sec. \ref{subsubsec:constructive}, Marchand {\em et al.} \cite{marchand1990convergence} added neurons to split the input space into pure regions containing only samples of the same class.  Liou {\em et al.}
\cite{liou1995ambiguous} proposed a three-layer network. Neurons in the
first layer represent cutting hyperplanes.  Their outputs are
represented as binary codes because of the use of threshold activation.
Each region corresponds to a vertex of a multidimensional cube. Neurons
in the second layer represent another set of hyperplanes cutting through
the cube. They split the cube into multiple parts again. Each part may
contain multiple vertices, corresponding to a set of regions in the
input space.  The output of the second layer is vertices in a
multidimensional cube.  The output layer offers another set of cuts.
Vertices at the same side are assigned the same class, which implies the
mergence of multiple regions.  The weights and biases of links can be
determined by BP.  Neurons are added if the samples cannot be well
differentiated based on the binary code at the layer output, which is
done layer by layer. 
Cabrelli {\em et al.} \cite{cabrelli2000constructive} proposed a two-layer network for convex
recursive deletion (CoRD) regions based on threshold activation.  They also interpreted hidden layers as separating hyperplanes that split the input space into multiple regions.  For each neuron, a region evaluates 0 or 1 depending on the side of the hyperplane that it lies in.
Geometrically, the data points in each region lie in a vertex of a
multidimensional cube. Their method does not require BP to update
weights. 

Results in \cite{marchand1990convergence, liou1995ambiguous,
cabrelli2000constructive} rely on threshold neurons.  One clear drawback
of threshold neurons is that they only have 0 or 1 binary outputs. A lot
of information is lost if the input has a response range (rather than a
binary value). FF-MLP uses the ReLU activation and continuous responses
can be well preserved. Clearly, FF-MLP is much more general.  Second, if
a sample point is farther away from the boundary, its response is
larger.  We should pay attention to the side of the partitioning
hyperplane a sample lies as well as the distance of a sample from the
hyperplane.  Generally, the response value carrys the information of the
point position.  This affects prediction confidence and accuracy.
Preservation of continuous responses helps boost the classification
accuracy.  

We should point out one critical reason for the simplicity of the FF-MLP
design. That is, we are not concerned with separating classes but
Gaussian blobs in the first layer. Let us take the 9-Gaussian-blobs
belonging to 3 classes as an example. There are actually $3^9$ ways to
assign 9 Gaussian blobs to three classes. If we are concerned with class
separation, it will be very complicated. Instead, we isolate each
Gaussian blob through layers $l_1$ and $l_2$ and leave the blob-class
association to the output layer. This strategy greatly simplifies our
design. 

\subsection{Relationship with LDA and SVM}\label{subsec:lda_relationship}

An LDA system can be viewed as a perceptron with specific weight and
bias as explained in Sec. \ref{sec:lda}. The main difference between an
LDA system and a perceptron is the method to obtain their model
parameters. While the model parameters of an LDA system are determined
analytically from the training data distribution, the parameters of MLP
are usually learned via BP. 

There is a common ground between FF-MLP and the support-vector machine
(SVM) classifier. That is, in its basic form, an SVM constructs the maximum-margin hyperplane to separate two classes, where the two-class
LDA plays a role as well. Here, we would like to emphasize their differences.
\begin{itemize}
\item SVM contains only one-stage while FF-MLP contains multiple stages
in cascade. Since there is no sign confusion problem with SVM, no ReLU
activation is needed. For nonlinear SVM, nonlinearity comes from
nonlinear kernels. It is essential for FF-MLP to have ReLU to filter out
negative responses to avoid the sign confusion problem due to
multi-stage cascade. 
\item The multi-class SVM is built upon the integration
of multiple two-class SVMs. Thus, its complexity grows quickly with the
number of underlying classes. In contrast, FF-MLP partitions Gaussian
blobs using LDA in the first stage. FF-MLP connects each Gaussian blob
to its own class type in Stage 3. 
\end{itemize}

We can use a simple example to illustrate the second item. If there are
$G$ Gaussian blobs belonging to $C$ classes, we have $C^G$ ways in
definining the blob-class association. All of them can be easily solved
by a single FF-MLP with slightly modification of the link weights in
Stage 3.  In contrast, each association demands one SVM solution. We
need $C^G$ SVM solutions in total.

\subsection{Relationship with Interpretable Feedforward CNN}\label{subsec:FF_CNN}

Kuo {\em et al.} gave an interpretation to convolutional neural networks
(CNNs) and proposed a feedforward design in \cite{kuo2019interpretable},
where all CNN parameters can be obtained in one-pass feedforward fashion
without back-propagation.  The convolutional layers of CNNs were
interpreted as a sequence of spatial-spectral signal transforms. The
Saak and the Saab transforms were introduced in \cite{kuo2018data} and
\cite{kuo2019interpretable}, respectively, for this purpose. The
fully-connected (FC) layers of CNNs were viewed as the cascade of
multiple linear least squared regressors.  The work of interpretable
CNNs was built upon the mathematical foundation in
\cite{kuo2016understanding, kuo2017cnn}. Recently, Kuo {\em et al.}
developed new machine learning theory called ``successive subspace
learning (SSL)" and applied it to a few applications such as image
classification \cite{chen2020pixelhop} \cite{chen2020pixelhop++}, 3D
point cloud classification \cite{zhang2020pointhop},
\cite{zhang2020pointhop++}, face gender classification
\cite{rouhsedaghat2020facehop}, etc. 

Although the FC layers of CNNs can be viewed as an MLP, we should point
out one difference between CNN/FC layers and classical MLPs.  The input
to CNNs is raw data (e.g. images) or near-raw data (e.g., spectrogram in
audio processing). Convolutional layers of CNNs play the role of feature
extraction. The output from the last convolutional layer is a
high-dimensional feature vector and it serves as the input to FC layers
for decision. Typically, the neuron numbers in FC layers are
monotonically decreasing. In contrast, the input to classical MLPs is a
feature vector of lower dimension. The neuron number of intermediate
layers may go up and down. It is not as regular as the CNN/FC layers. 

\section{Conclusion and Future Work}\label{sec:conclusion}

We made an explicit connection between the two-class LDA and the MLP
design and proposed a general MLP architecture that contains two
intermediate layers, denoted by $l_1$ and $l_2$, in this work. The
design consists of three stages: 1) stage 1: from input layer $l_{in}$
to $l_1$, 2) stage 2: from intermediate layer $l_1$ to $l_2$, 3) stage 3: from
intermediate layer $l_2$ to $l_{out}$. The purpose of stage 1 is to partition
the whole input space flexibly into a few half-subspace pairs. The
intersection of these half-subspaces yields many regions of interest.
The objective of stage 2 is to isolate each region of interest from
other regions of the input space. Finally, we connect each region of
interest to its associated class based on training data. We use Gaussian
blobs to illustrate regions of interest with simple examples. In
practice, we can leverage GMM to approximate datasets of general shapes.
The proposed MLP design can determine the MLP architecture (i.e. the
number of neurons in layers $l_1$ to $l_2$) and weights of all links in
a feedforward one-pass manner without any trial and error. Experiments
were conducted extensively to compare the performance of FF-MLP and the
traditional BP-MLP. 

There are several possible research directions for future extension.
First, it is worthwhile to develop a systematic pruning algorithm so as
to reduce the number of partitioning hyperplanes in stage 1. Our current
pruning method is a greedy search algorithm, where the one that has the
least impact on the classification performance is deleted first. It does
not guarantee the global optimality. 
Second, it is well known that BP-MLP is vulnerable to
adversarial attacks. It is important to check whether FF-MLP encounters
the same problem. 
Third, we did not observe any advantage of using the
backpropagation optimization in the designed FF-MLP in Sec.
\ref{sec:experiments}. One conjecture is that the size of the proposed
FF-MLP is too compact. The BP may help if the network size is larger.
This conjecture waits for further investigation. 
Fourth, for a general sample distribution, we approximate the distribution with Gaussian blobs that may have different covariances. This does not meet the homoscedasticity assumption of the two-class LDA. It is possible to improve the system by replacing LDA with heteroscedastic variants.

\section*{Acknowledgment}

This material is based on research sponsored by US Army Research
Laboratory (ARL) under contract number W911NF2020157. The U.S.
Government is authorized to reproduce and distribute reprints for
Governmental purposes notwithstanding any copyright notation thereon.
The views and conclusions contained herein are those of the authors and
should not be interpreted as necessarily representing the official
policies or endorsements, either expressed or implied, of US Army
Research Laboratory (ARL) or the U.S. Government. Computation for the
work was supported by the University of Southern California's Center for
High Performance Computing (hpc.usc.edu). 

\bibliographystyle{IEEEtran}
\bibliography{mybibfile}

\end{document}